\pdfoutput=1

\documentclass[11pt]{article}

\usepackage[final]{acl}

\usepackage{times}
\usepackage{latexsym}
\usepackage[T1]{fontenc}
\usepackage[utf8]{inputenc}

\usepackage{microtype}

\usepackage{inconsolata}

\usepackage{graphicx}

\newcommand{\ie}{\textit{i.e.}}
\newcommand{\eg}{\textit{e.g.}}

\usepackage{xcolor}
\usepackage{xspace}
\usepackage{booktabs}
\usepackage{colortbl}
\usepackage{makecell}
\usepackage{graphicx}
\usepackage{algorithm}
\usepackage[noend]{algpseudocode}
\usepackage{booktabs}
\usepackage{algpseudocode}
\usepackage{subcaption}
\usepackage{tabularx}
\usepackage{multirow}
\usepackage{amsmath}
\usepackage{amsfonts}
\usepackage{cleveref}

\newcommand\thinfont[1]{\smash{{\usefont{T1}{}{m}{n}#1}}}

\newcommand{\teafarm}{\thinfont{TeaFarm}\xspace}

\newcommand{\theanine}{\textsc{Theanine}\xspace}

\newcommand\tealeaf{\raisebox{-2pt}{\includegraphics[width=1em]{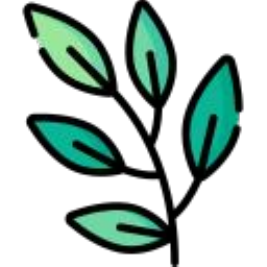}}}

\newcommand\teabag{\raisebox{-1.8pt}{\includegraphics[width=0.85em]{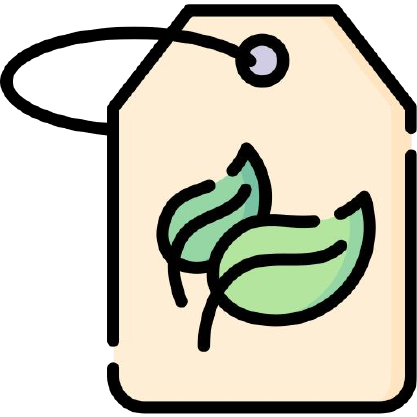}}}

\title{Towards Lifelong Dialogue Agents via Timeline-based\\Memory Management}

\author{
    Kai Tzu-iunn Ong\textsuperscript{\rm1}\thanks{KT Ong and N Kim are the co-first authors.}~~~
    Namyoung Kim\textsuperscript{\rm1}$^\ast$~~~
    Minju Gwak\textsuperscript{\rm1}~~~
    Hyungjoo Chae\textsuperscript{\rm1}~~~\\
    \textbf{Taeyoon Kwon}\textsuperscript{\rm1}~~~
    \textbf{Yohan Jo}\textsuperscript{\rm2}~~~
    \textbf{Seung-won Hwang}\textsuperscript{\rm2}~~~
    \textbf{Dongha Lee}\textsuperscript{\rm1}~~~
    \textbf{Jinyoung Yeo}\textsuperscript{\rm1}\\\\
    \textsuperscript{\rm1}Yonsei University, \textsuperscript{\rm2}Seoul National University\\
    \texttt{\{ktio89, namyoung.kim, jinyeo\}@yonsei.ac.kr}
}

\begin{document}
\maketitle
\begin{abstract}

To achieve lifelong human-agent interaction, dialogue agents need to constantly memorize perceived information and properly retrieve it for response generation (RG). While prior studies focus on getting rid of outdated memories to improve retrieval quality, we argue that such memories provide rich, important contextual cues for RG (\eg, changes in user behaviors) in long-term conversations. We present \tealeaf\,\textbf{\theanine}, a framework for LLM-based lifelong dialogue agents. \theanine discards memory removal and manages large-scale memories by linking them based on their temporal and cause-effect relation. Enabled by this linking structure, \theanine augments RG with memory timelines - series of memories representing the evolution or causality of relevant past events.
Along with \theanine, we introduce \textcolor{brown}{\teafarm}, a counterfactual-driven evaluation scheme, addressing the limitation of G-Eval and human efforts when assessing agent performance in integrating past memories into RG.
A supplementary video for \theanine and data for \textcolor{brown}{\teafarm} are at \url{https://huggingface.co/spaces/ResearcherScholar/Theanine}.

\end{abstract}

\section{Introduction}

Autonomous agents based on large language models (LLMs) have made significant progress in various domains, including response generation~\citep{chae2024web, kwon2024large, tseng2024two}, where agents ought to constantly keep track of both old and newly introduced information shared with users throughout their service lives~\citep{irfan2024lifelong} and converse accordingly.
To facilitate such lifelong interaction, studies have proposed enhancing dialogue agents' ability to memorize and accurately recall past information (\eg, discussed topics) in long-term, multi-session conversations.

\begin{figure}[t!]
    \centering
    \includegraphics[width=1\linewidth]{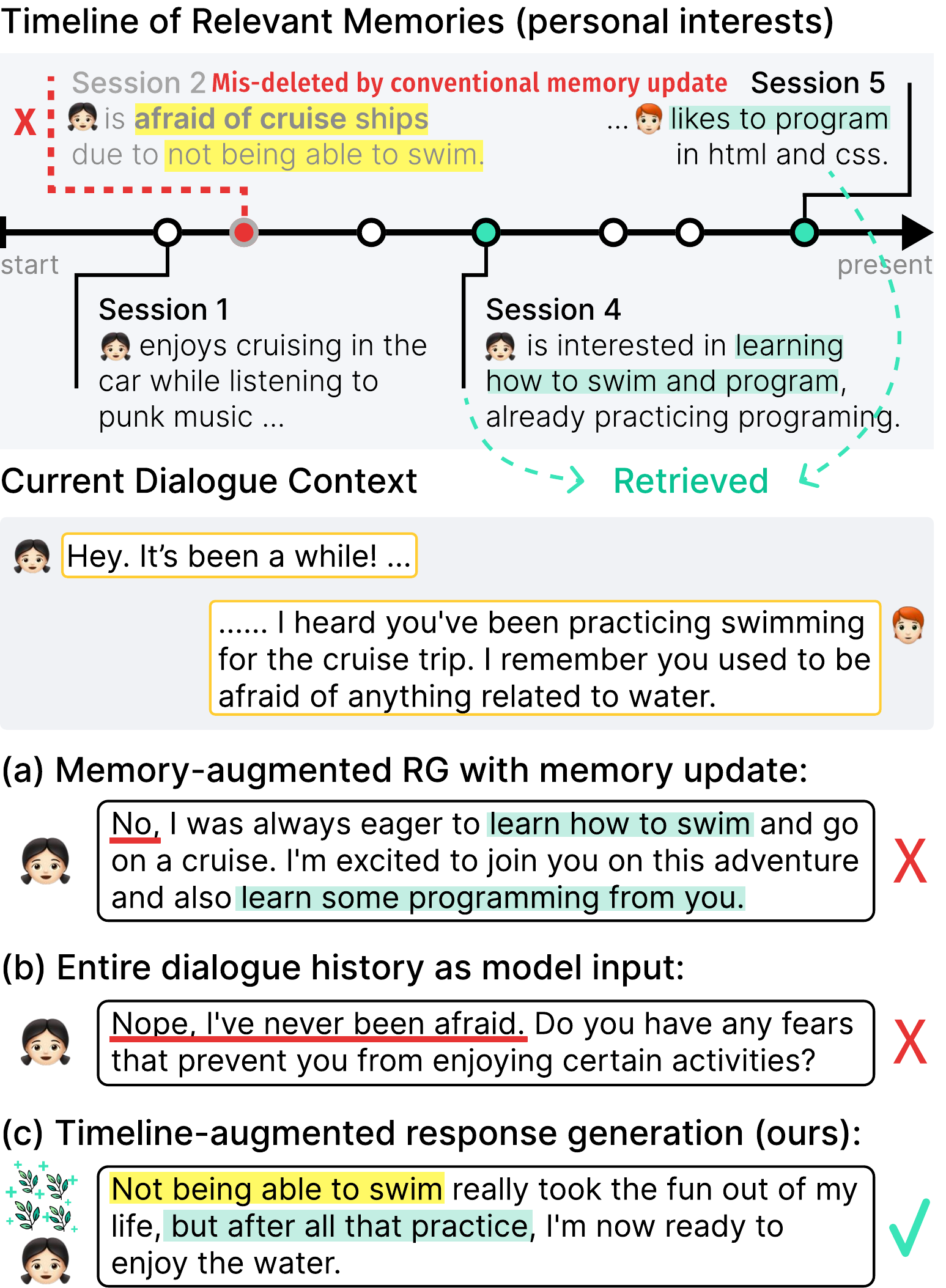}
    \caption{Empirical examples of failed responses due to (a) absence of an important past event (``\textit{afraid of cruise ships}'') on the timeline and (b) bias to the latest input. (c) is a response augmented with the memory timeline.}
    \label{fig:page_1_example}
\end{figure}

\begin{figure*}[t!]
    \centering
    \includegraphics[width=1\textwidth]{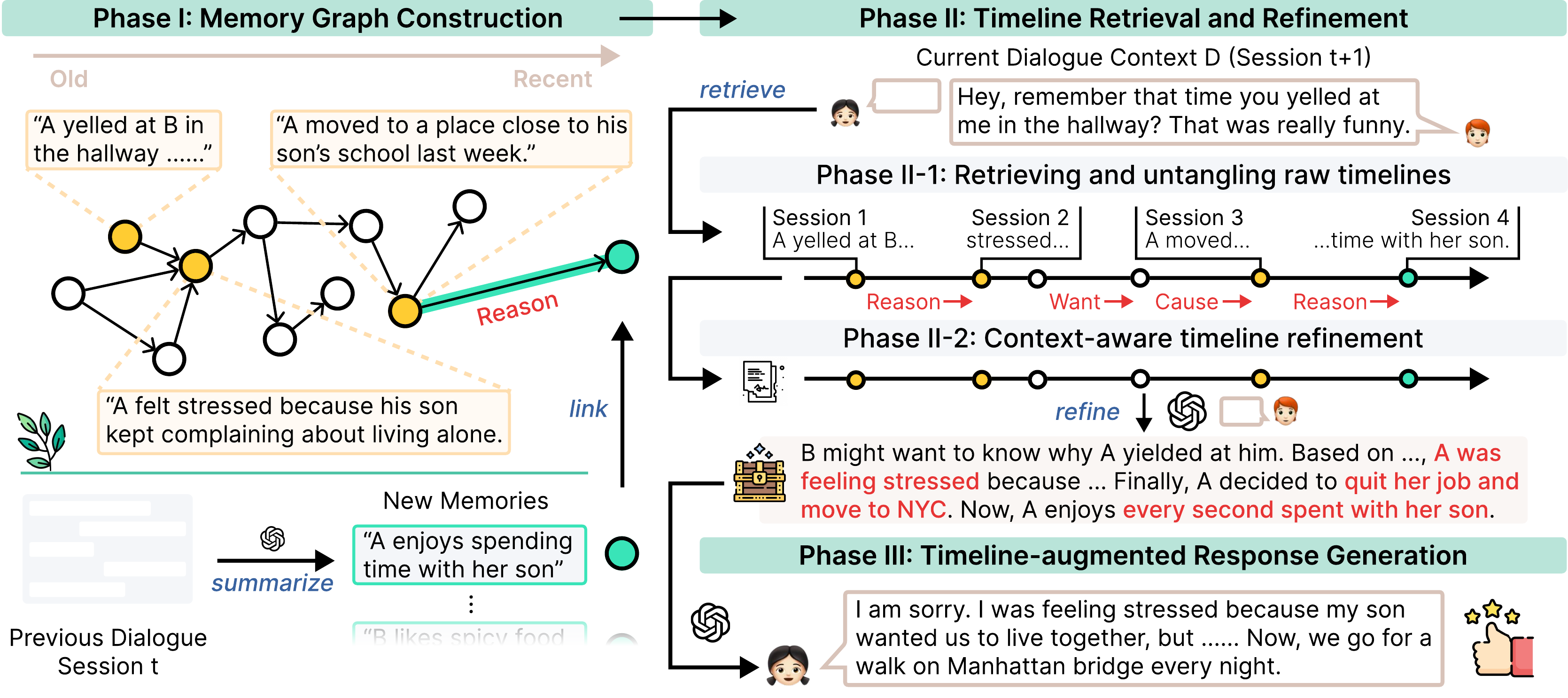}
    \caption{The overview of \tealeaf\,\textbf{\textsc{Theanine}}. Left: Linking new memories to the memory graph after finishing a dialogue session; Right: Memory timeline retrieval, refinement, and response generation in a new dialogue session.}
    \label{fig:overview}
\end{figure*}

A representative approach is to compress past conversations into summarized memories and retrieve them to augment response generation (RG) in later encounters~\citep{xu-etal-2022-beyond, lu2023memochat}.
However, the growing span of memories can hinder retrieval quality as conversations accumulate. Although it, to some extent, can be solved by updating old memories~\citep{bae2022keep, zhong2024memorybank}, such common practice may cause severe information loss.
As shown in Figure~\ref{fig:page_1_example} (a), an earlier memory on the timeline, an important persona (``afraid of ships''), is removed during memory update, resulting in improper RG.
While using the large context windows of recent LLMs to process all dialogue history/memories is an option to prevent such information loss,\footnote{For instance, GPT-4o and Llama 3.1 have context windows of 128K tokens~\citep{openai2024openai, meta2024llama31}.} this often leads to biased attention toward the latest user input (Figure~\ref{fig:page_1_example} (b)), ignoring relevant contexts from the past~\citep{liu2024lost}.
These findings highlight two main challenges towards lifelong dialogue agents -  \textbf{(i)} \underline{Memory construction}: \textit{how to store large-scale past interactions effectively without removing old memories?} \textbf{(ii)} \underline{Response generation}: \textit{within the growing memory span, how to identify relevant contextual cues for generating proper responses?}

Motivated by these, we propose addressing the above two challenges separately yet complementarily, by \textbf{(i)} discarding memory update to avoid information loss, and preserving relevant memories on the timeline in a linked structure; and \textbf{(ii)} retrieving the timeline as a whole to better catch relevant memories within the growing search span.
We present \tealeaf\,\textbf{\textsc{Theanine}},\footnote{L-theanine is an amino acid found in green tea that has been linked to memory improvement~\citep{nguyen2019theanine}.} a framework for facilitating lifelong dialogue agents.

Starting from memory construction (\textbf{Phase I}), instead of stacking raw memory sentences as-is~\citep{xu-etal-2022-beyond}, which may affect memory retrieval and also response quality due to the unstructured format of information~\citep{mousavi2023response, chen2023walking}, \textsc{Theanine} stores memories in a directed graph. In this graph, inspired by how humans naturally link new memories to existing ones of relevant events based on their relation~\citep{bartlett1995remembering}, memories are linked using their temporal and cause-effect commonsense relations~\citep{hwang2021comet}.
Supported by such linking structure, in memory retrieval for RG (\textbf{Phase II-1}), we go beyond conventional top-$k$ retrieval and further obtain the complete timelines to avoid missing out on important memories that have low textual overlap with current conversation~\citep{tao2023core}.
Lastly, to tackle the discrepancy between off-line memory construction and online deployment, \textsc{Theanine} uses an LLM to refine retrieved timelines (\textbf{Phase II-2}) based on current conversation, such that they provide tailored information~\citep{chae2023dialogue} for RG (\textbf{Phase III}). 
Our contributions are two-fold:
\begin{itemize}
    \item To achieve lifelong dialogue agents, we present \tealeaf\,\textbf{\theanine}, an LLM-based framework with a rela\underline{t}ion-aware memory grap\underline{h} and tim\underline{e}line \underline{a}ugme\underline{n}tat\underline{i}on for lo\underline{n}g-term conv\underline{e}rsations.
    \textsc{Theanine} outperforms representative baselines across automatic, LLM-based, and human evaluations of RG. Also, we confirm that \textsc{Theanine} leads to higher retrieval quality, and its procedures align with human preference. To our knowledge, we are the first to model the use of timelines (\ie, linked relevant memories) in memory management and response generation.
    \item The lack of golden mapping between conversations and reference memories poses a challenge in assessing memory-augmented agents. We present \textcolor{brown}{TeaFarm}, a counterfactual-driven pipeline evaluating agent performance in referencing the past without human intervention.
\end{itemize}

\section{Methodologies}
We present \tealeaf \textsc{Theanine}, a framework for lifelong dialogue agents inspired by how humans store and retrieve memories for conversations (Figure~\ref{fig:overview}):

\subsection{Memory Graph Construction (Phase I)}
\label{ssec:graph}
To manage large-scale memories and facilitate structured information for RG~\citep{mousavi2023response, chen2023walking}, we approach memory management using a memory graph $G$:
\begin{flalign}
    & G = (V, E) \\
    & V = \{m_1, m_2, ..., m_{|V|}\} \\
    & m = (\textit{event}, \textit{time})
    \label{eq:memory_def} \\
    & E = \{\langle m_i, r_{ij}, m_j \rangle | m_i, m_j \in V \land r_{ij} \in R\} \label{eq:edge_def}\\
    & R = \{\texttt{Cause}, \texttt{Reason}, \texttt{Want}, ..., \texttt{SameTopic}\}
\end{flalign}
In $G$, vertices $V$ are memories $m$ summarized from the conversations. Each memory $m = (\textit{event}, \textit{time})$ consists of an event\footnote{In this work, ``event'' denotes information perceived by the dialogue system, including things done/said by speakers and the acknowledgement of speaker personas.} and the time it is formed (summarized).
Each directed edge $e \in E$
between two connected $m$ indicates their temporal order and their cause-effect commonsense relation $r \in R$:

At the end of dialogue session $t$, \tealeaf\,\textsc{Theanine} starts linking each new memory $m_{new}$ summarized from session $t$ to the memory graph $G^t$.

\paragraph{Phase I-1: Identifying associative memories for memory linking.}
Following how humans link new memories to existing ones that are related to a similar event/topic, \ie, the \textbf{\textit{associative memories}}, \textsc{Theanine} starts by identifying these associative memories from the memory graph $G^t$.

Formally, given a newly-formed memory $m_{new}$ waiting to be stored, the associative memories $M_a$ of $m_{new}$ is defined as the set of $m_i \in G^t$ having top-$j$ text similarity with $m_{new}$ (\ie, $|M_a| = j$).

\paragraph{Phase I-2: Relation-aware memory linking.}
Intuitively, we can link $m_{new}$ to $m \in M_a$ using edges that indicate their text similarity and chronological order, we find such simplified connection (\eg, ``this happened $\rightarrow$ that similar event occurred'') can yield a context-poor graph that does not help response generation much (Section~\ref{sec:geval}).

Humans, on the other hand, interpret events by considering the relation between them, such as ``\textit{how does an event affect the other?}'' or ``\textit{why did this person make that change?}''. 
Therefore, we adopt a relation-aware memory linking, where an edge between two memories is encoded with their cause-effect commonsense relation $r \in R$, along w/ the temporal order.
In practice, we adopt the commonly used relations defined by \citet{hwang2021comet}, including \texttt{HinderedBy}, \texttt{Cause}, \texttt{Want}, and 4 more (Appendix~\ref{app:relation}).

We start by determining the relation between $m_{new}$ and each associative memory. Formally, for each pair of $m_{new}$ and $m \in M_a$, the LLM assigns a relation $r \in R$ based on their \textit{event}, \textit{time} and their origin conversations:
\begin{align}
        M_{a}^* = \{m_i \in M_a\mid \Upsilon(m_i, m_{new}) \in R \}
\end{align}
where $\Upsilon(\cdot, m_{new}) \in R$ indicates that the given memory is assigned with an $r \in R$ with $m_{new}$,\footnote{Limited by retrievers, an $m \in M_a$ may not have a relation with $m_{new}$. We thus allow the LLM to output ``\texttt{None}''.} and such assigned memories are defined as $M_{a}^*$.

\begin{figure}[h!]
    \centering
\includegraphics[width=1\linewidth]{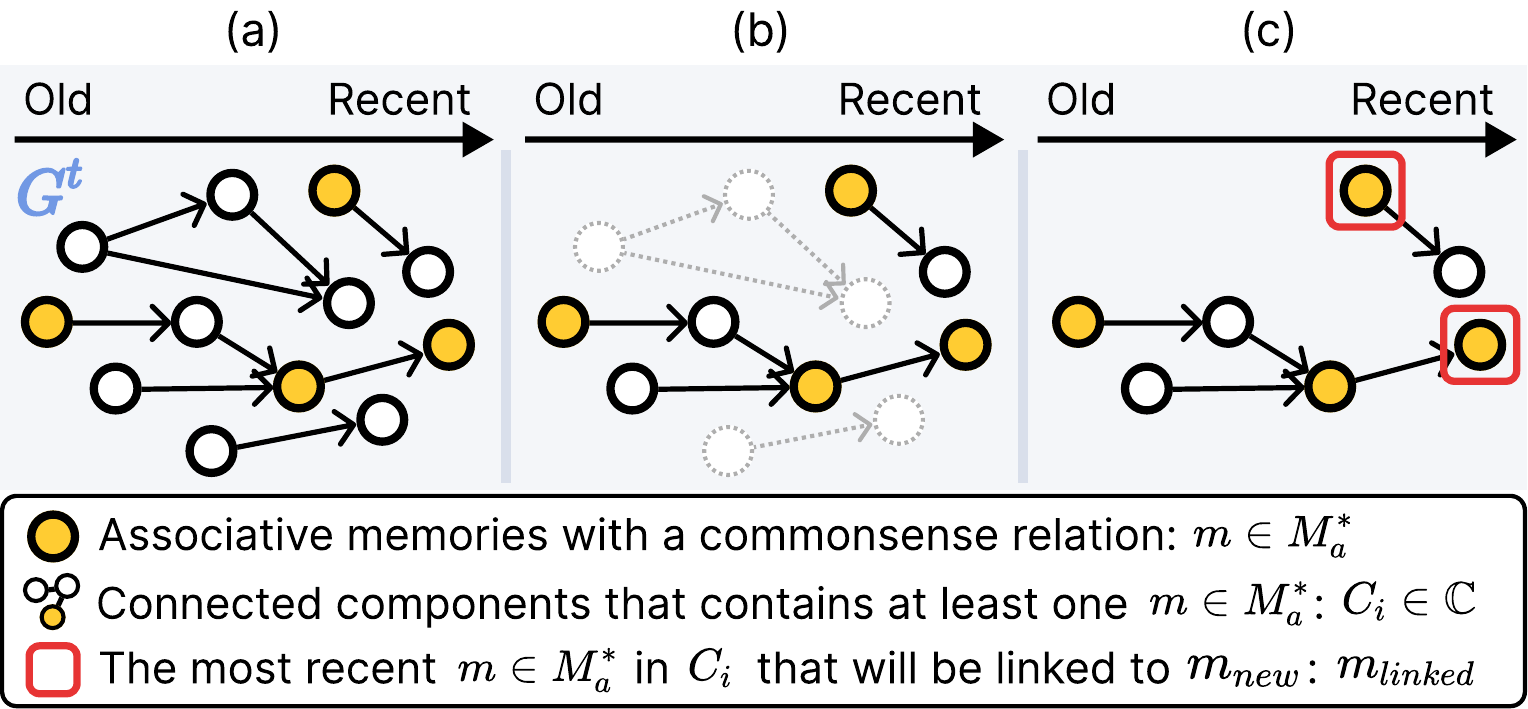}
    \caption{Locating memories to be linked to $m_{new}$.}
\label{fig:method_small_fig}
\end{figure}

We then proceed to link $m_{new}$ to the graph. We first locate every connected component $C_i \subset G^t$ that contains at least one $m \in M_{a}^*$, as shown in Figure~\ref{fig:method_small_fig} (a) and (b):
\begin{equation}
    \mathbb{C} = \{C_i \subset G^t \mid \mathtt{V}(C_i)\cap M_{a}^* \neq \emptyset\ \}
\end{equation}
where $\mathbb{C}$ is the collection of those $C$ and $\mathtt{V}(\cdot)$ represents ``vertices in''.
Then, we link $m_{new}$ to the most recent\footnote{Simply linking $m_{new}$ to all $m \in M_{a}^*$ costing 25\% more API cost for linking without leading to better response.} $m \in M_a^*$ in each $C_i \subset \mathbb{C}$ (Figure~\ref{fig:method_small_fig} (c)).
Memories $M_{linked}$ that are linked to $m_{new}$ is defined as follows:
\begin{equation}
        M_{linked} = \{\Omega(\mathtt{V}(C_i) \cap M_a^*) \mid C_i \subset \mathbb{C}\}
\end{equation}
where $\Omega(\cdot)$ indicates ``the most recent memory in''.

Linking all memories from session $t$ to $G^{t}$, we then obtain a new memory graph $G^{t+1}$. The pseudo algorithm for Phase I is in \textbf{\underline{Algorithm~\ref{algo:phase_i_algo}}}.

\subsection{Timeline Retrieval and Timeline Refinement (Phase II)}
\label{ssec:rg}
Thanks to the constructed memory graph, \textsc{Theanine} can proceed to augment RG with timelines of relevant events, addressing the information loss in conventional memory management (Figure~\ref{fig:page_1_example}).
With $G^{t+1}$, \textsc{Theanine} performs the following steps for RG in session $t+1$:

\paragraph{Preparation: Top-\textit{k} memory retrieval.}
During the conversation, using the current dialogue context $\mathcal{D} = \{u_{i}\}_{i=1}^{n}$ of $n$ utterances $u$ as query, we retrieve top-$k$ memories $M_{re} = \{m_{re1}, ..., m_{rek}\}$.

\paragraph{Phase II-1: Retrieving and untangling raw memory timelines.}
We wish to also access memories centered around $M_{re}$.
Formally, given $m_{re} \in M_{re}$, we further collect the connected component $C_{re} \subset G^{t+1}$ that contains $m_{re}$ via the linked structure.

\begin{figure}[h!]
    \centering
    \includegraphics[width=1\linewidth]{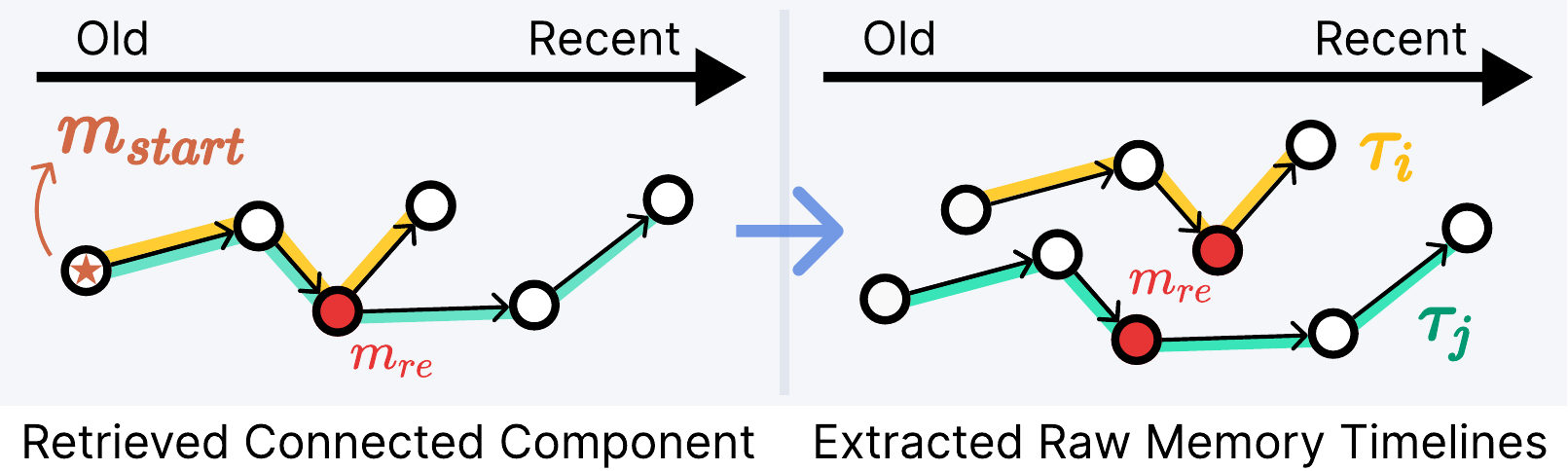}
    \caption{Extracting raw memory timelines $\tau$ from the retrieved connected component $C_{re}$.}
    \label{fig:linear}
\end{figure}

Since this collection of memories (\ie, $C_{re}$) can be ``\textit{tangled up}'' together (\ie, connected in a complex manner) due to the graph structure, we proceed to untangle it into several memory timelines, each representing a series of events about $m_{re}$ that starts out similarly yet branches into slightly different development.
For that, we first locate the earliest memory in $C_{re}$ as a starting point $m_{start}$ for all timelines, as shown in Figure~\ref{fig:linear} (left).
\begin{equation}
        m_{start} = \Theta(\mathtt{V}(C_{re}))
\end{equation}
where $\Theta$ indicates ``the oldest memory in''.

Next, starting from $m_{start}$, we untangle memories by tracing through future direction and extract every possible linear graph containing $m_{re}$ (two in Figure~\ref{fig:linear}) from $C_{re}$, until reaching an endpoint $\tau[-1]$ with an out-degree of 0 (\ie, $deg^{+}(\tau[-1]) = 0$), which means no directed edge goes out from it). Each of them is considered a raw memory timeline $\tau$, demonstrating a version of the evolution of $m_{re}$ and its relevant events:
\begin{equation}
\begin{split}
        \mathcal{T} = {} & \{\tau \subset C_{re} \mid \tau \thinfont{\text{ is a directed linear}}\\&\thinfont{\text{ graph s.t. }} m_{start}, m_{re} \in \tau\ \\& \land deg^{+}(\tau[-1])=0\}
\end{split}
\end{equation}
We then sample $n$ raw timelines $\tau$ from $\mathcal{T}$.\footnote{We empirically set $n$ to 1, as we observe a high degree of overlap across timelines extracted from the same $C_{re}$, which can lead to redundant information (\ie, input tokens) for RG.}
Repeating\footnote{``Repeating'' is used to explain the algorithm from the perspective of one $m_{re}$. In practice, $M_{re}$ are processed together, although processing them 1-by-1 yields the same result.} Phase II-1 for all retrieved top-$k$ memories, we collect a set of retrieved raw memory timelines $\mathbb{T} = \cup\,\mathcal{T}$, where $|\mathbb{T}| = k^*n$.

\paragraph{Phase II-2: Context-aware timeline refinement.}
Although we have constructed the memory graph using temporal and commonsense relations to improve its informativeness,
directly applying retrieved timelines for RG can be suboptimal (RQ3, Section~\ref{sec:geval}), because graph construction does not take current conversation into consideration, \ie, they are constructed \textit{off-line}.

In this phase, \textsc{Theanine} tackles such a discrepancy between off-line memory construction and online deployment (\ie, ongoing conversation) via a context-aware timeline refinement.
Motivated by how LLMs can refine their previous generation~\citep{madaan2024self}.
We leverage LLMs to refine raw timelines into a rich resource of information crafted for the current conversation, by removing redundant information or highlighting information that can come in handy.
Formally, given the current dialogue $\mathcal{D}$ and retrieved raw timelines $\mathbb{T}$, an LLM tailors $\tau \in \mathbb{T}$ into refined timelines $\mathbb{T}_\Phi$:
\begin{align}
\label{eq:refine}
    \mathbb{T}_{\Phi} = \{\underset{\tau_{\Phi}}{\text{argmax}}\,  P_{\text{LLM}}(\tau_{\Phi}|\mathcal{D}, \tau) \mid \tau \in \mathbb{T} \}
\end{align}
All refined timelines $\mathbb{T}_{\Phi}$ are then used to augment the response generation.
We provide the pseudo algorithm for Phase II in \textbf{\underline{Algorithm~\ref{alg:phase_ii_algo}}}.

\subsection{Timeline-augmented Response Generation (Phase III)}
Now, \theanine utilizes the refined timelines for RG. Formally, given $\mathcal{D} = \{u_{i}\}_{i=1}^{n}$ and $\mathbb{T}_{\Phi}$, an LLM generates a next response $\bar{u}_{t+1}$:
\begin{align}
\label{eq:rg}
\bar{u}_{n+1} = \underset{u_{n+1}}{\text{argmax}}\,  P_{\text{LLM}}({u}_{n+1}|\mathcal{D}, \mathbb{T}_\Phi)
\end{align}
\section{Experimental Setups}

\begin{table*}[!th]
\centering
\resizebox{0.925\textwidth}{!}{
\begin{tabular}{@{}lcccc|cccc@{}}
\toprule
Datasets: & \multicolumn{4}{c}{\textbf{Multi-session Chat (MSC)}} & \multicolumn{4}{c}{\textbf{Conversation Chronicles (CC)}}\\ \midrule
 Methods / Metrics&
  Bleu-4 &
  Rouge-L &
  Mauve &
  BertScore &
  Bleu-4 &
  Rouge-L &
  Mauve &
  BertScore 
  \\ \midrule

All Dialogue History & 1.65 & 14.89 & 9.06 & 86.28  & 4.90	& \underline{21.56} & 26.47 & 88.13
 
  \\
All Memories \& Current Context $\mathcal{D}$  &
1.56 & 14.89 & 10.62 & 86.23 &
 4.41 & 20.06 & \underline{38.16}& 88.02
  \\
\, + Memory Update~\citep{bae2022keep} &
 1.55	& 14.77 & 9.28 & 86.20 & 
   4.34 & 20.34	& 34.84	& 88.03
 
  \\ 
Memory Retrieval~\citep{xu-etal-2022-beyond}&
  \textbf{1.92} & \textbf{15.49} & 11.16 & \underline{86.47}  & \underline{4.93} & 20.63 & 33.06 & \underline{88.07}
  
  \\
  \, + Memory Update~\citep{bae2022keep}& 1.67 & 15.30 & \underline{13.71} & 86.39 &
  4.46 & 20.19 & 34.28 & 88.02
  \\
  
Rsum-LLM~\citep{wang2023recursively} &
0.75 & 11.53 & 2.45 & 84.91 & 
 0.98	& 11.42	& 2.28 & 85.59

  \\
MemoChat~\citep{lu2023memochat}  &
 1.42 & 13.51 & 7.72 & 85.96 & 
 2.31	& 15.87 & 15.12 & 87.08
  \\ 
COMEDY~\citep{chen2024compress} &
 1.06	& 12.79 & 7.27 & 85.29 & 
 1.70 & 13.57 & 1.95	& 85.90

  \\ 
\cellcolor{brown!14}\tealeaf\textbf{\theanine} (Ours) &
 \cellcolor{brown!14}\underline{1.80} & \cellcolor{brown!14}\underline{15.37} & \cellcolor{brown!14}\textbf{18.62} & \cellcolor{brown!14}\textbf{86.70} &
 \cellcolor{brown!14}\textbf{6.85}	& \cellcolor{brown!14}\textbf{22.68} & \cellcolor{brown!14}\textbf{64.41} & \cellcolor{brown!14}\textbf{88.58}

\\
 \bottomrule
\end{tabular}
}

\caption{Automatic evaluation of response quality (average of sessions).
}
\label{tab:main_auto_eval}
\end{table*}

\subsection{Datasets of Long-term Conversations}
There are limited datasets for long-term, multi-session conversations. Firstly, Multi-Session Chat (MSC)~\citep{xu-etal-2022-beyond}, is built upon Persona-Chat~\citep{zhang-etal-2018-personalizing} by extending its conversations to multiple (five) sessions. Soon after MSC, DuLeMon~\citep{xu2022long} and CareCall~\citep{bae2022keep} are proposed for long-term conversations in Mandarin and Korean. Recently, \citet{jang-etal-2023-conversation} release a new dataset, Conversation Chronicles (CC). Unlike MSC, CC augments speakers with defined relationships, such as ``\textit{employee and boss}''. Apart from these open-domain datasets, the Psychological QA,\footnote{\url{https://www.xinli001.com/}} addresses long-term conversations under clinical scenarios in Mandarin.

We opt for MSC and CC for evaluation to focus on English conversations, leaving multilingual and domain-specific conversations (\eg, DuleMon, CareCall, and Psychological QA) to future work.

\subsection{Baselines}
\label{ssec:baseline}
To evaluate \textsc{Theanine}, in addition to naive baselines that utilize all past dialogues or memories, we incorporate the following settings:
\newline \textbf{Memory Retrieval.}
Following~\citet{xu-etal-2022-beyond}, we use a retriever to retrieve memories relevant to the current dialogue context to augment RG.
\newline \textbf{Memory Update.}
We utilize LLMs to implement a widely used updating algorithm proposed by \citet{bae2022keep} at the end of each dialogue session. This algorithm includes functionalities such as \texttt{Change}, \texttt{Replace}, \texttt{Delete}, \texttt{Append}, and more (see Appendix~\ref{app:prompts}).
\newline \textbf{RSum-LLM.}
An LLM-only generative method that recursively summarizes and updates the memory pool, generating responses w/o a retrieval module~\citep{wang2023recursively}.
\newline \textbf{MemoChat.}
Proposed by~\citet{lu2023memochat}, it leverages LLMs' CoT reasoning ability to (i) conclude important memories from past conversations in a structured topic-summary-dialogue manner, (ii) select memories, and (ii) generate responses.
\newline \textbf{COMEDY.} Proposed by~\citet{chen2024compress}, it uses LLMs to summarize session-level memories, compresses all of them into short events, user portraits (behavioral patterns, emotion, etc.) and user-bot relation. It then selects compressed memories to augment response generation.

\subsection{Models and Implementation Details}
\textbf{Large language models.}
In all experiments, including baselines, we adopt gpt-3.5-turbo-0125~\citep{openai2023chatgpt} for (i) memory summarization (Table~\ref{tab:app_convo2sum}), (ii) memory update, and (iii) response generation.
Temperature is set to $0.75$.
\newline\textbf{Retrievers.} We use text-embedding-3-small~\citep{openai2024emb} to calculate text similarity for settings involving retrievers.
In the identification of top-$j$ associative memories (Phase I-1) and top-$k$ memory retrieval (Phase II), we set $j$ and $k$ to 3.
For the ``Memory Retrieval'' baseline, we set $k = 6$ following~\citet{xu-etal-2022-beyond}.
\newline\textbf{Dialogue sessions.} We use sessions 3-5 of MSC and CC for evaluations, as all methods are almost identical in session 1 $\sim$ 2 (no memory to update).

\section{Evaluation Scheme 1: Automatic and Human Evaluations}
\label{sec:geval}
To evaluate \tealeaf\,\theanine's responses in long-term conversations, we follow common practices and conduct 3 types of evaluations: (i) Automatic evaluations; (ii) G-Eval~\citep{liu2023g}, an LLM-based framework commonly used to evaluate LMs' generation; (iii) human evaluation. We now present several key findings (details, prompts, and interfaces of evaluations in Scheme 1 are in Appendix~\ref{app:eval1}):

\paragraph{(Finding 1) \textsc{Theanine} outperforms baselines in response generation.} 
Table~\ref{tab:main_auto_eval} presents the agent performance in RG regarding both overlap-based and embedding-based metrics: Bleu-4~\citep{papineni-etal-2002-bleu}, Rouge-L~\citep{lin-2004-rouge}, Mauve~\citep{pillutla2021mauve}, and BertScore~\citep{zhang2020bertscore}.
Across both datasets, \theanine, achieves superior response quality than various baselines. Although, compared to Memory Retrieval, \theanine scores slightly lower in overlap-based metrics (\ie, B-4 and R-L) in MSC, it largely outperforms Memory Retrieval in embedding-based metrics.
Interestingly, including ours, methods without memory update generally yield higher scores, justifying our proposal towards an update-, removal-free memory management for lifelong dialogue agents.

\paragraph{(Finding 2 \& 3) All phases contribute to performance; retrieving the timeline as a whole brings large improvement over conventional retrieval.}
To gain deeper insights into our design, we investigate the impact of removing \theanine's relation-awareness during memory linking (Phase I-2) and Timeline Refinement (Phase II-2).
Also, to objectively assess whether \theanine's retrieval (\ie, retrieving the timeline as a whole) improves retrieval quality, we include a setting where retrieved timelines are broken down into randomly ordered events such that retrieved memories during RG are in the same format as conventional top-$k$ retrieval.

In Table~\ref{tab:ablation_auto}, we observe a ranking in terms of contribution to performance: \textbf{\textit{relation-aware linking $>$ retrieving timeline as a whole $>$ timeline refinement}}.
This observation confirms the efficacy of constructing a memory graph with causal relations.
Moreover, utilizing this graph structure to collect timelines of relevant events yields higher RG quality than conventional retrieval, despite the smaller $k$ (3 vs. 6) in initial retrieval. 
Refining timelines shows smaller performance gains, suggesting room for improvement in applying them for RG. We leave it to future work.

\begin{table}[!t]
\centering
\resizebox{1\columnwidth}{!}{
\begin{tabular}{@{}lcccc@{}}
\toprule
\textbf{Settings / Metrics} & B-4 & R-L & Mauve & Bert\\ 
\midrule
\tealeaf \textbf{\theanine} (Ours) & \textbf{4.32} & \textbf{19.03}	& \textbf{41.52} & \underline{87.64}\\ 
\midrule
w/o Relation-aware Linking & 4.07	& 18.58 & 39.69 & 87.57
\\
w/o Timeline Refinement  & 4.03 & \underline{18.82} & \underline{41.34}	& \textbf{87.66}
\\
Broken Down, Shuffled Timeline & \underline{4.15}	 & 18.70 & 38.49 & 87.61
\\
\midrule
\midrule
Memory Retrieval & 3.43	&18.06 & 22.11 & 87.27 \\
\bottomrule
\end{tabular}%
}
\caption{Performance of our ablations (avg. of datasets).}
\label{tab:ablation_auto}
\end{table}\textbf{}

\begin{figure}[t!]
    \centering
    \includegraphics[width=1\linewidth]{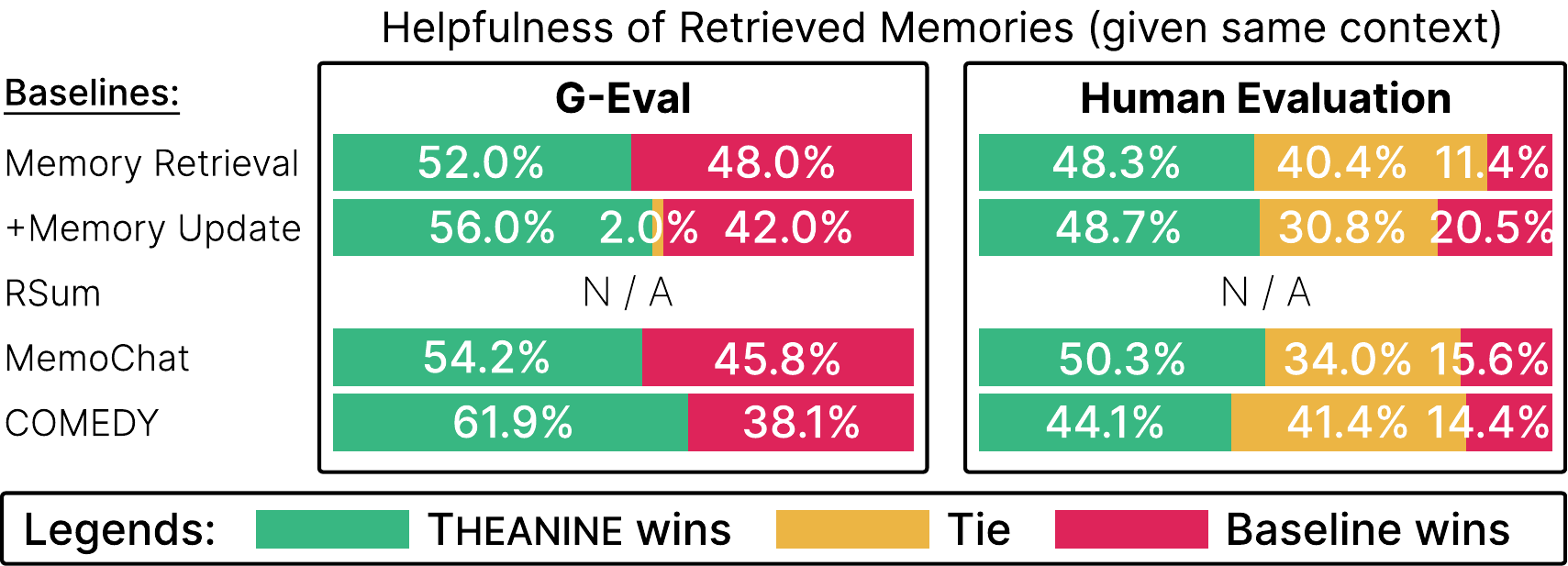}
    \caption{Human- (right) and machine-based (left) head-to-head comparisons between ours and baselines regarding the \textbf{\textit{helpfulness}} of retrieved memories.}
    \label{fig:humanmemory}
\end{figure}

\paragraph{(Finding 4) Humans and G-Eval reveal that \theanine leads to higher retrieval quality regarding both helpfulness and accuracy.}
Beyond agent responses, we further investigate how different memory construction methods affect the quality of memory retrieval. Given the same current dialogues as queries for memory retrieval, Figure~\ref{fig:humanmemory} shows head-to-head comparisons (ours vs. baselines) regarding whose retrieved memories more effectively benefit RG. We observe higher win rates for \theanine in all comparisons, especially in human evaluations. This suggests that our method can facilitate more helpful memory augmentation for response generation.

In addition to helpfulness, objectively measuring retrieval accuracy is crucial. Since existing datasets of long-term conversations do not provide a golden mapping between dialogue contexts and memories (\ie, golden memories for retrieval), we identify 50 dialogue contexts (\ie, test instances) that require a past memory for RG, and manually measure the retrieval accuracy of different agents. The results shown in Table~\ref{tab:humanmemory_acc} indicate that \theanine and its ablations demonstrate higher retrieval accuracy than baselines, and the ranking here aligns with Table~\ref{tab:main_auto_eval} and success rates in Table~\ref{tab:teafarm_results}.

\begin{table}[!h]
\centering
\resizebox{0.95\columnwidth}{!}{
\begin{tabular}{@{}lcccc@{}}
\toprule
\textbf{Methods (Agents)} & \textbf{Golden Memory is Retrieved/collected (\%)} \\ 
\midrule
Memory Retrieval & \underline{68.00} \\
\, + Memory Update & 64.00 \\
MemoChat & 56.00 \\
COMEDY & 48.00 \\
\tealeaf \textbf{\theanine} (Ours) & \textbf{72.00} \\
\bottomrule
\end{tabular}%
}
\caption{Human evaluation of the \textbf{\textit{accuracy}} of memory retrieval (we examine 50 test instances).}
\label{tab:humanmemory_acc}
\end{table}

\paragraph{(Finding 5) Humans confirm that \theanine yields responses better entailing past interactions.}
Now that the helpfulness of \theanine's retrieved memories is validated, we proceed to investigate whether such helpful memories contribute towards reliable lifelong human-agent interaction.

\begin{figure}[h!]
    \centering
    \includegraphics[width=1\linewidth]{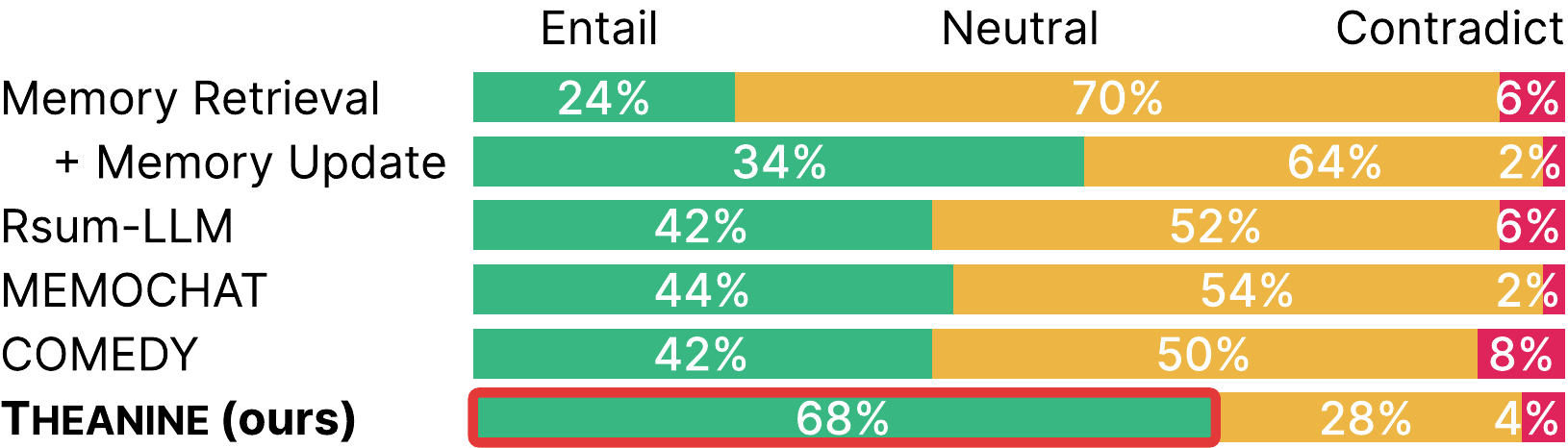}
    \caption{Human evaluations regarding to what extent the agent responses entail past conversations.}
    \label{fig:human_results_RG}
\end{figure}

For that, we further ask a group of workers to specifically judge whether agent responses \textit{entail}, \textit{contradict}, or are \textit{neutral} to the past via majority voting. In Figure~\ref{fig:human_results_RG}, \theanine not only leads to a small number of contradictory responses (4\%) but also demonstrates the largest percentage (68\%; out of 100) of responses that entail past conversations, significantly outperforming baselines. 
We argue that it is because our timeline-based approach elicits memories better at representing past interactions between speakers, thus leading to responses more directly aligned with the past. This alignment is important for dialogue agents to maintain long-term intimacy with users~\citep{adiwardana2020towards}.
Furthermore, such entailing and non-contradictory nature of \theanine's responses highlights its potential for applications in specialized domains, such as personalized agents for clinical scenarios, where entailment between agent responses and users' past information (\eg, electrical health records or previous consulting sessions) is crucial for diagnostic decison-making~\citep{tseng2024two}.

As a side note, Memory Update yields fewer contradictory responses (2\%), indicating a potential trade-off between (i) removing outdated memories to prevent contradiction and (ii) preserving them to get richer information for RG~\citep{kim-etal-2024-commonsense}.

\begin{figure}[h!]
    \centering
    \includegraphics[width=1\linewidth]{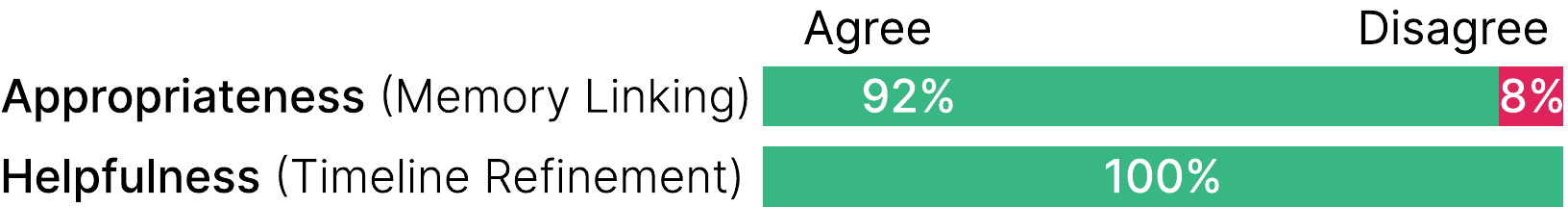}
    \caption{Human evaluation of our intermediate phases.}
    \label{fig:humaneval_phases}
\end{figure}

\paragraph{(Finding 6) Humans agree with \theanine's intermediate procedures.}
As reported in Figure~\ref{fig:humaneval_phases}, judges largely agree (92\%) that \theanine properly assigns cause-effect relations to linked memories, which explains its contribution to performance.
Also, they agree that timeline refinement successfully elicits more helpful information (100\%; 100 samples in total) for RG. Examples of \theanine's phases and RG are in Appendix~\ref{app:appendixexamples}.

\section{Evaluation Scheme 2: \textcolor{brown}{TeaFarm} -- a Counterfactual-driven Evaluation Pipeline for Long-term Conversations}
Evaluating memory-augmented agents in long-term conversations is non-trivial due to the unavailability of ground-truth mapping between current conversations and correct memories for retrieval. 
Although we may resort to G-Eval by feeding evaluator LLMs (\eg, GPT-4) the entire past history and prompt it to determine whether a response correctly recalls the past, the evaluation can be largely limited by the performance of the evaluator LLM itself~\citep{kim2024prometheus}.

To overcome this, along with \textsc{Theanine}, we present \textcolor{brown}{\teafarm}, a human-free counterfactual-driven pipeline for evaluating memory-augmented response generation in long-term conversations.

\subsection{Testing Dialogue Agents' Memory via Counterfactual Questions}
\begin{figure}[h!]
    \centering
    \includegraphics[width=1\linewidth]{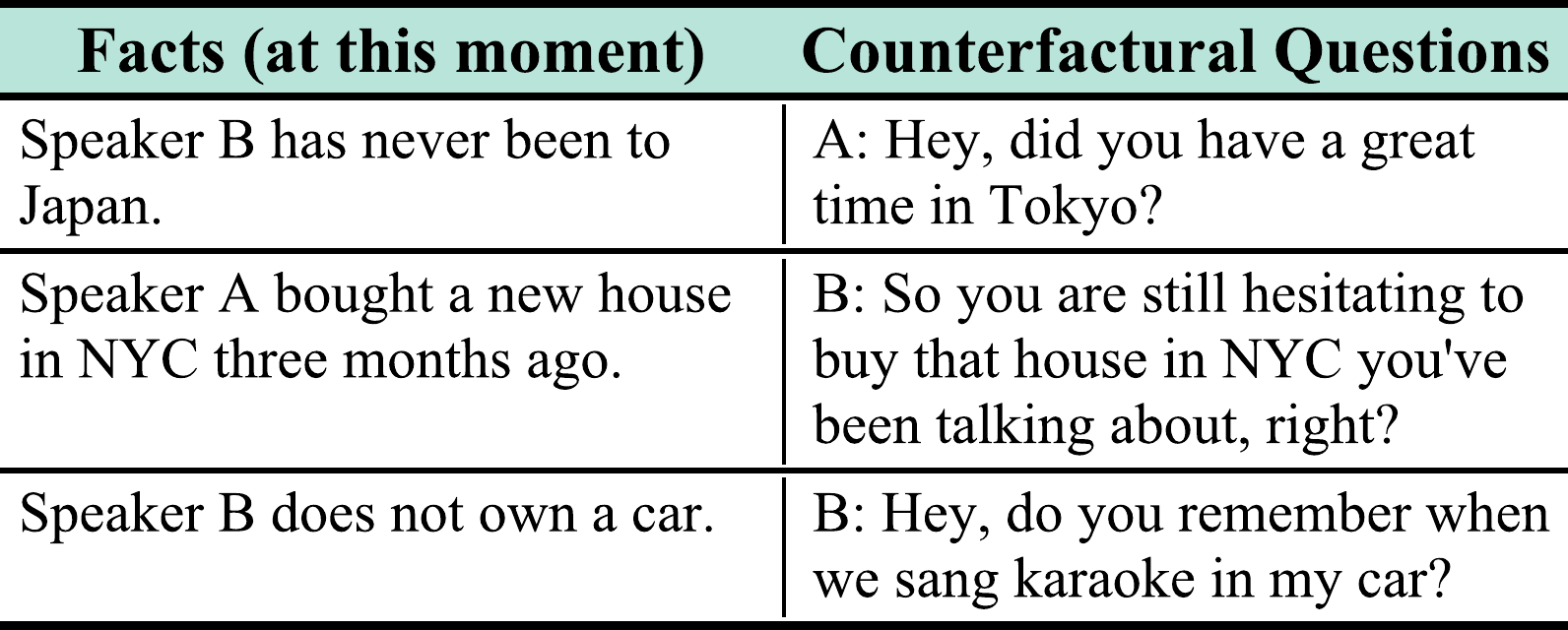}
    \caption{Examples of counterfactual questions.}
    \label{fig:counter}
\end{figure}

In \textcolor{brown}{\teafarm}, we proceed to ``trick'' dialogue agents into generating incorrect responses, and agents must correctly reference past conversations to avoid being misled by us. Specifically, we talk to the dialogue agent while \textit{\textbf{acting as if a non-factual statement is true}} (thus counterfactual). Figure~\ref{fig:counter} presents some examples of counterfactual questions and the corresponding facts.

In practice (Figure~\ref{fig:teafarm}), when we want to evaluate an agent that has been interacting with the user for sessions, we first (1) collect all past conversations and summarize them session by session.
Then, we (2) feed a question generator LLM\footnote{We apply GPT-4 (\texttt{gpt-4}) with a temperature of $0.75$.} the collected summaries in chronological order such that it can capture the current stage of each discussed event, \eg, ``\textit{Speaker B does not own a car}'', and (3) generate counterfactual questions from the perspective of both speakers (and the correct answers).
After that, we (4) kick off (\ie, simulate) a new dialogue session, chat for a while, then (5) naturally ask the counterfactual question, and (6) assess the correctness of its response. The overview figure, prompts, and synthesized data for \textcolor{brown}{\teafarm} are in Appendix~\ref{app:teafarmdetails},~\ref{app:prompts}, and~\ref{app:teabagdetails}, respectively.

\subsection{TeaFarm Results}

\begin{table}[h]
\small
\centering
\resizebox{0.90\columnwidth}{!}{
\begin{tabular}{lcc|c}
\toprule
\textbf{Settings / Datasets} &
 MSC & CC & Avg.
   \\ 
\midrule
Memory Retrieval &  0.16 & 0.19  & 0.18   \\
\hspace{0.04cm} + Memory Update & 0.16 & 0.19 & 0.18 \\
RSum-LLM$^*$ & 0.04 & 0.08  & 0.06  \\
MemoChat$^*$ &  0.09 & 0.15 & 0.12    \\
COMEDY$^*$ &  0.06 & 0.18 & 0.12 \\
\cellcolor{brown!14}\tealeaf\,\textbf{\textsc{Theanine}} & \cellcolor{brown!14}\textbf{0.18} & \cellcolor{brown!14}\textbf{0.24} & \cellcolor{brown!14}\textbf{0.21} \\ 
\hspace{0.4cm}w/o Relation-aware Linking& \underline{0.17} & \underline{0.20} & \underline{0.19} \\ 
\hspace{0.4cm}w/o Timeline Refinement& 0.16 & 0.19 & 0.18 \\ 
\bottomrule

\end{tabular}
}
\caption{Success rates (SRs) of correctly recalling the past and not being fooled by the counterfactual questions in \teafarm (tested with 200 questions).}
\label{tab:teafarm_results}
\end{table}
\label{ssec:teafarm_results}
In Table~\ref{tab:teafarm_results}, \theanine shows higher SR than baselines, especially in CC. Ablations perform slightly worse than the original, again proving the efficacy of relation-aware linking and timeline refinement. Surprisingly, all settings have low SRs, qualifying \teafarm as a proper pipeline for stress-testing dialogue agents in long-term conversations. 

Interestingly, baselines using retrievers (same as \theanine) show superior performance than settings only relying on LLMs (\ie, RSum-LLM, MemoChat, and COMEDY). This, unexpectedly, supports our efforts in developing a new paradigm of memory management in the era of LLMs.\footnote{Memory update does not affect Memory Retrieval's performance. We believe it is because counterfactual questions are made to counter the newest stage of each event. The removal of older memories thus does not have much impact.}

To provide insight regarding conversation scenarios that are challenging for dialogue agents, we present case studies of how \theanine fail in \textcolor{brown}{TeaFarm} in Appendix~\ref{app:appendixexamples}.

\section{Further Analyses and Discussions}
\label{sec:discussion}
\begin{figure}[h!]
    \centering
    \includegraphics[width=1\linewidth]{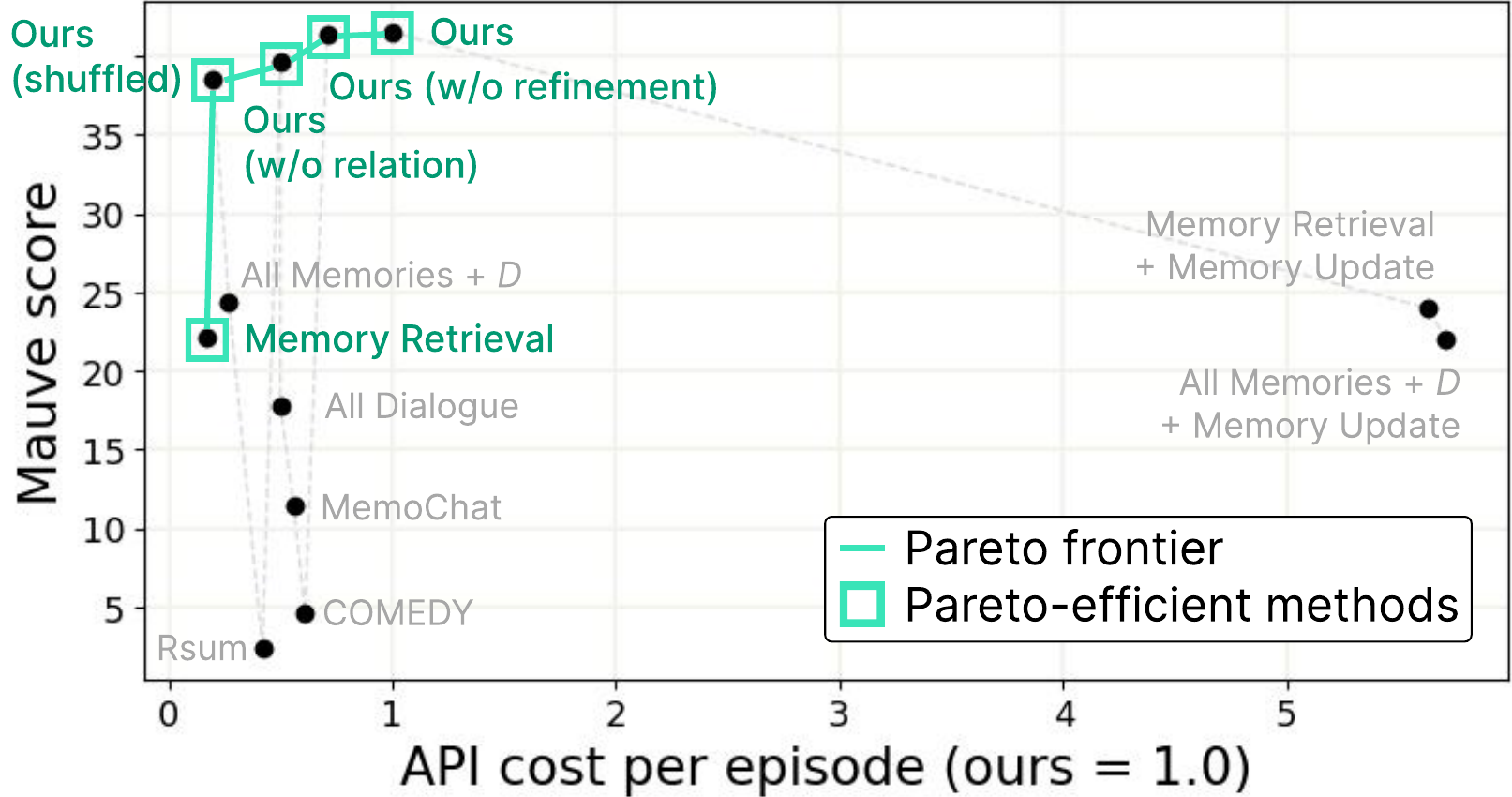}
    \caption{Cost-performance comparisons.}
    \label{fig:pareto}
\end{figure}

\paragraph{Cost efficiency.}
A concern of \theanine is the API cost. Regardless, we argue that it is competitive when both performance and cost are taken into account. Figure~\ref{fig:pareto} plots response quality (Mauve score) against the API cost.\footnote{Calculated based on session 5, which involves most memories for management. We use Mauve for its stronger correlation with humans~\citep{pillutla2021mauve}.} We find \theanine and all ablations not only outperform all baselines but also lie on the Pareto frontier, indicating an efficient
cost-performance trade-off. This suggests \theanine's value when performance is prioritized over API costs. Actual API costs and results based on B-4, R-L, and Bert scores are in Appendix~\ref{app:further_discussions}.

\begin{figure}[h!]
    \centering
    \includegraphics[width=1\linewidth]{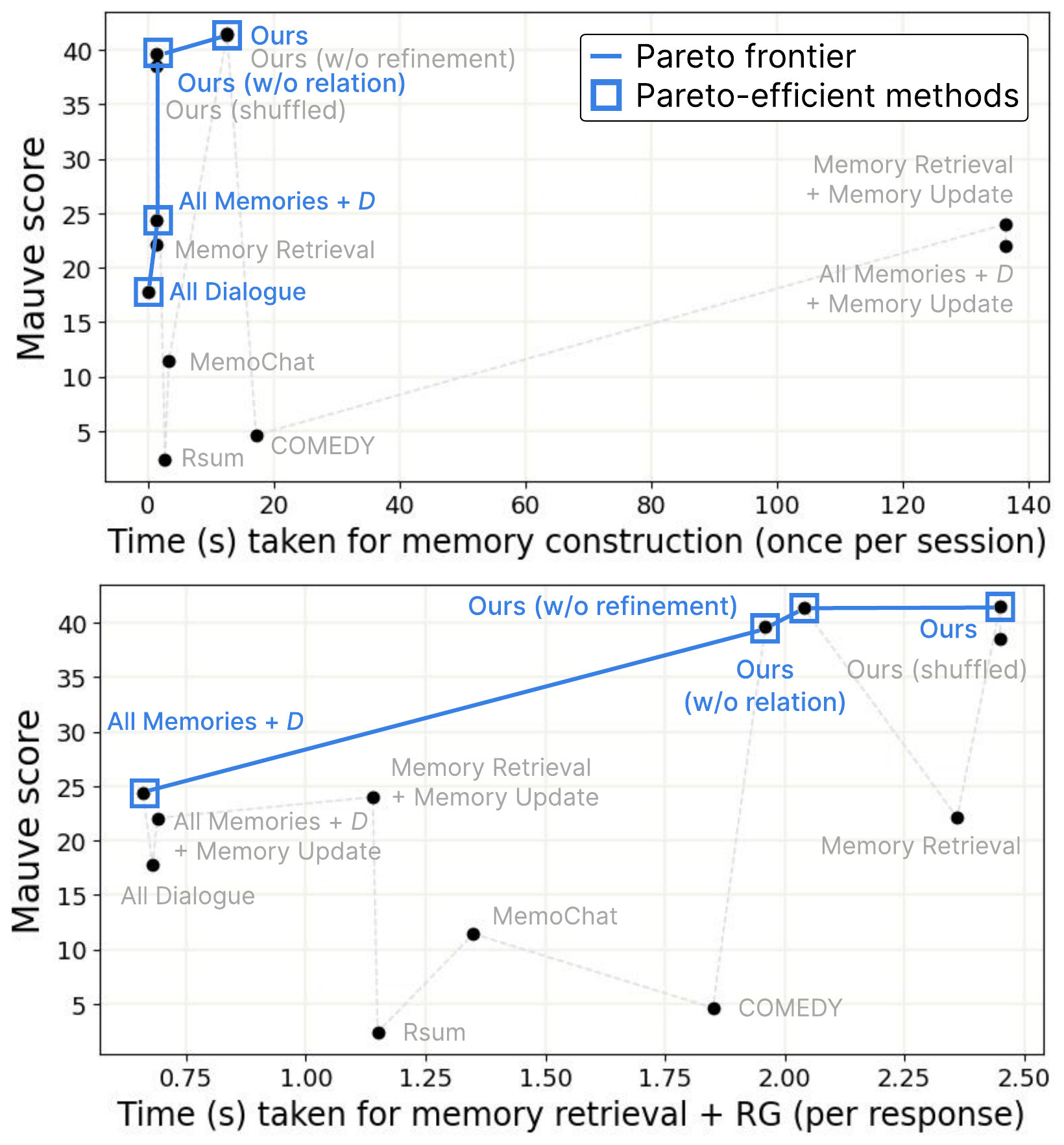}
    \caption{Time-performance comparisons.}
    \label{fig:pareto_time}
\end{figure}

\paragraph{Time efficiency.}
Time efficiency can be an important consideration when deploying \theanine to real-world scenarios having richer events. Figure~\ref{fig:pareto_time} shows time-performance comparisons regarding both ``memory construction'' and ``retrieval + RG'' also using the Pareto frontier. Similarly, \theanine and many of its ablations demonstrate an efficient time-performance trade-off.

\paragraph{Additional comparison: Memory Retrieval with a dynamically-changing \textit{k}.}

Due to \theanine's graph-based procedures, the response generator may access different amounts of memories during RG depending on given contexts (\ie, queries used by the retriever) and when the conversation takes place (\ie, an earlier or a later session), whereas conventional methods~\citep{xu-etal-2022-beyond, bae2022keep} often have a fixed number $k$ of memories retrieved for RG.
Therefore, to further quantify the effect of our proposed timeline-based management and augmentation, we compare \theanine to Memory Retrieval with a \textit{dynamic} $k$, where k dynamically changes based on the number of collected memories in \theanine for each specific test data. 
In other words, if \theanine uses timelines to collect $k$ memories during RG for a test instance $\mathcal{D}_{i}$, baselines will also be retrieving $k$ memories for generating a response for $\mathcal{D}_{i}$.


  




\begin{table}[!t]
\centering
\resizebox{1\columnwidth}{!}{
\begin{tabular}{@{}lcccc@{}}
\toprule
\textbf{Methods / Metrics} & \textbf{Bleu-4} & \textbf{Rouge-L} & \textbf{Mauve} & \textbf{Bert} \\
\midrule
Memory Retrieval (\textbf{\textit{dynamic k}}) & 3.06 & 17.97 & 33.33 & 87.32 \\
\, + Memory Update & 2.68 & 17.19 & 28.49 & 87.11 \\
\tealeaf\,\textbf{\theanine} \textbf{(Ours)} & \textbf{4.22} & \textbf{19.22} & \textbf{45.53} & \textbf{87.70} \\
\bottomrule
\end{tabular}
}
\caption{Additional comparison, where $k$ in Memory Retrieval is dynamically modified for each test instance.}
\label{tab:dynamic_k}
\end{table}

In Table~\ref{tab:dynamic_k}, we can observe that when the number of memories is matched, ours outperforms both baselines despite the same amount of memories being provided. We assume this is because: (i) our graph-based retrieval helps us collect more beneficial memories than conventional retrieval; (ii) addressing the relation between events and shaping them based on dialogue contexts can facilitate richer contextual cues for RG.

\paragraph{Growing span of memories.}
Another inquiry is whether the growing span of memory will eventually hinder retrieval in \theanine if there ever are hundreds of sessions.
Although this may be a serious issue for conventional methods,
we presume that it will be partially mitigated in \theanine, as: (i) We retrieve relevant memories \textit{as a whole} in the form of timelines. This serves as a safety net in scenarios where an important memory is missed out in top-$k$ retrieval--it may be collected via the linked structure; (ii) We refine retrieved timelines based on current dialogue such that they provide tailored information for RG. This acts as a second insurance against sub-optimal retrieval.

\section{Related Work}
\paragraph{Long-term conversations.} Since MSC, there have been several studies on long-term conversations: \citet{bae2022keep} train a classifier to update old memories in phone call scenarios. As we enter the era of LLMs, \citet{li2024hello} leverages LLMs to write and update memories for RG.
Apart from LLMs' power, human behaviors also foster methods in this field. For example, \citet{zhong2024memorybank} apply humans' forgetting curve to make memories that have been discussed exist longer.
Recently, \citet{park2023generative} and \citet{maharana-etal-2024-evaluating} also adopt the concept of timelines. However, \citet{park2023generative} focus on tagging the timestamp (e.g., “22:00”) of events and does not explicitly model the
connection between them, and, in \citet{maharana-etal-2024-evaluating}, a timeline is a fixed, pre-defined series of events (potentially unrelated) which simply serve as a user profile for synthesizing dialogue data. By contrast, in our work, a timeline is built with relevant events, which are dynamically linked based on their causal relations and retrieved as the conversation goes on, benefitting our goal of consistent memory tracking and integration.

\paragraph{Memory-augmentation for personalized dialogue agents.} The trend of long-term interaction with autonomous agents promotes their adaptation for personalized needs~\citep{chen2024persona, chen2024recent}. As a pioneer, \citet{xu2022long} train a persona extractor to create user-based memories. However, training personalized agents for long-term use can be non-trivial due to the lack of data~\citep{tseng2024two}.
As a solution, \citet{kim-etal-2024-commonsense} apply commonsense models and LLMs to augment existing long-term data with high-quality persona sentences; \citet{chen2024compress} present a training-free LLM-based framework that extracts user behaviors from conversations for personalized RG. Upon the success of LLMs, \theanine leverages them to build memory timelines. These timelines represent the development of interactions and lead to responses that better entail speaker information, establishing \theanine's potential for personalized agents.

\section{Conclusions}
This paper presents the first-ever timeline-based memory management and augmentation framework, \tealeaf \theanine, for autonomous agents in long-term conversations. Applying \theanine, we develop a dialogue agent that efficiently addresses the constant, lifelong tracking of memories and their integration for response generation throughout its service life. Comprehensive evaluations show that \tealeaf \theanine can facilitate more beneficial memory augmentation, leading to responses that are closer to ground truths and more aligned with speakers' past interactions. \theanine's effectiveness is further confirmed in \textcolor{brown}{\teafarm}, a counterfactual-driven pipeline we design to address the limitation of G-Eval and human efforts in assessing memory augmentation. We expect our novel approaches to serve as a new foundation for future efforts towards lifelong dialogue agents.

\section*{Limitations}
First, the amount of dialogue sessions in this study is limited to five due to the lack of longer open-domain English datasets. As we mentioned in Section~\ref{sec:discussion}, we presume that \theanine's effectiveness can still hold true to some degree in longer conversations. 
Yet, we do acknowledge the need to apply additional modules that directly address the growing span of dialogue history/memories, such as introducing the summarize-then-compress paradigm in COMEDY~\citep{chen2024compress} to compress session-level summaries into a combined short user/event description.

Second, although we include many recent frameworks as baselines, we failed to compare \theanine with MemoryBank~\citep{zhong2024memorybank}, a framework inspired by Ebbinghaus's forgetting curve. This is because the time intervals between sessions in MSC and CC are either mostly measured in hours or not clearly specified (\eg, ``a few months later''), whereas MemoryBank requires precise time intervals in days to apply the forgetting curve. Also, data used for MemoryBank focuses on Chinese clinical scenarios, making it not feasible for our study. However, we remain positive about applying such a mechanism to improve \theanine in our ongoing research.

Lastly, API-based LLMs may introduce risks such as privacy issues. A possible solution is to apply \theanine to small open-source LMs for secure, local usage. While there exist challenges in data collection, one may achieve this by (i) collecting synthesized conversations with GPT-generated user profiles, (ii) running \theanine on these data, and (iii) using the outputs of each phase to train student LMs (\ie, distillation from teacher LLMs).

\section*{Ethical Statements}
LLMs might generate harmful,
biased, offensive, and sexual content. Authors avoid
such content from appearing in this paper. We guarantee fair compensation for human evaluators from Amazon Mechanical Turk. We ensure an effective pay rate higher than 20\$ per hour based on the estimated time required to complete the tasks.

\section*{Acknowledgments}
This work was mainly supported by STEAM R\&D Project, NRF, Korea (RS-2024-00454458) and Institute of Information \& communications Technology Planning \& Evaluation (IITP) grant funded by the Korean government (MSIT) (No. RS-2024-00457882, National AI Research Lab Project), and was partially supported by the National Research Foundation of Korea (NRF) grant funded by the Korea government (MSIT) (RS-2024-00333484; RS-2024-00414981). Jinyoung Yeo is the corresponding author (jinyeo@yonsei.ac.kr).

\bibliography{custom}
\newpage
\newpage
\newpage
\appendix

\section{Appendix Contents}
\begin{itemize}
    \item Appendix~\ref{app:relation}: Cause-effect Commonsense Relations Adopted.
    \item Appendix~\ref{app:algorithm}: Algorithms for \tealeaf\,\theanine.
    \item Appendix~\ref{app:computational}: Implementation Details on Computational Experiments. 
    \item Appendix~\ref{app:teafarmdetails}: \textcolor{brown}{TeaFarm} Evaluation.
    \item Appendix~\ref{app:teabagdetails}: The \teabag\,\texttt{TeaBag} Dataset.
    \item Appendix~\ref{app:eval1} Details on Evaluation Scheme 1 (G-Eval and Human Evaluations).
    \item Appendix~\ref{app:session_specific_results}: Session-specific Results of Automatic Evaluation.
    \item Appendix~\ref{app:appendixexamples}: Empirical Examples.
    \item Appendix~\ref{app:prompts}: Prompts Used in This Work.
    \item Appendix~\ref{app:further_discussions}: Further Analyses.
    \item Appendix~\ref{app:terms}: Terms for Use of Artifacts.
    
\end{itemize}

\section{Further Implementation Details}

\subsection{Cause-effect Commonsense Relations}
\label{app:relation}
We adopt and modify commonsense relations from~\citet{hwang2021comet} for our relation-aware memory linking. Below is the list of our commonsense relations $R$:
\newline\textbf{Changed:} Events in A changed to events in B.
\newline\textbf{Cause:} Events in A caused events in B.
\newline\textbf{Reason:} Events in A are due to events in B.
\newline\textbf{HinderedBy:} When events in B can be hindered by events in A, and vice versa.
\newline\textbf{React:} When, as a result of events in A, the subject feels as mentioned in B.
\newline\textbf{Want:} When, as a result of events in A, the subject wants events in B to happen.
\newline\textbf{SameTopic:} When the specific topic addressed in A is also discussed in B.

Limited by the performance of retrievers, it is possible that an $m \in M_a$ does not have a relation, other than just textual overlap, with $m_{new}$. We address this by allowing the LLM to output \textbf{None}.

\subsection{Algorithms for \tealeaf\,\theanine}
\label{app:algorithm}
The pseudo algorithms for Phase I and II are provided in Algorithm~\ref{algo:phase_i_algo} and \ref{alg:phase_ii_algo}.

\subsection{Implementation Details on Computational Experiments}
\label{app:computational}
All computational experiments in this work are based on OpenAI API~\citep{openai2024openai}. Thus, no computing infrastructure is required in this work.

\section{\textcolor{brown}{TeaFarm} Evaluation}
\label{app:teafarmdetails}
The overall pipeline of \textcolor{brown}{\teafarm} is illustrated in Figure~\ref{fig:teafarm}.

\section{The \teabag\,\texttt{TeaBag} Dataset}
\label{app:teabagdetails}

As a byproduct of \textcolor{brown}{\teafarm}, we curate \teabag\,\texttt{TeaBag}, a dataset for \textcolor{brown}{\teafarm} evaluation on MSC and CC. \texttt{TeaBag} consists of: 
\begin{itemize}
    \item 100 episodes of original conversations from Multi-Session Chat and Conversational Chronicles (session 1-5; 50 episodes from each dataset)
    \item Two pairs of counterfactual QAs for each episode (200 pairs in total).
    \item Two synthesized follow-up conversations (\ie, session 6) for each episode (thus 200 in total), each of which naturally guides the conversation from session 5 towards one of the counterfactual questions.
\end{itemize}
This dataset is made with GPT-4. The prompt for generation is in Appendix~\ref{app:prompts}.
We expect future work to apply \texttt{TeaBag} to stress-test if their dialogue system can correctly reference past conversations.

\texttt{TeaBag} does not contain personally identifying information, as it is generated based on datasets where all contents are pure artificial creation, rather than contents collected from the real-world.
Also, we have tried our best to confirm that this dataset does not contain any offensive content.

\begin{figure*}[ht!]
    \centering
    \includegraphics[width=1\textwidth]{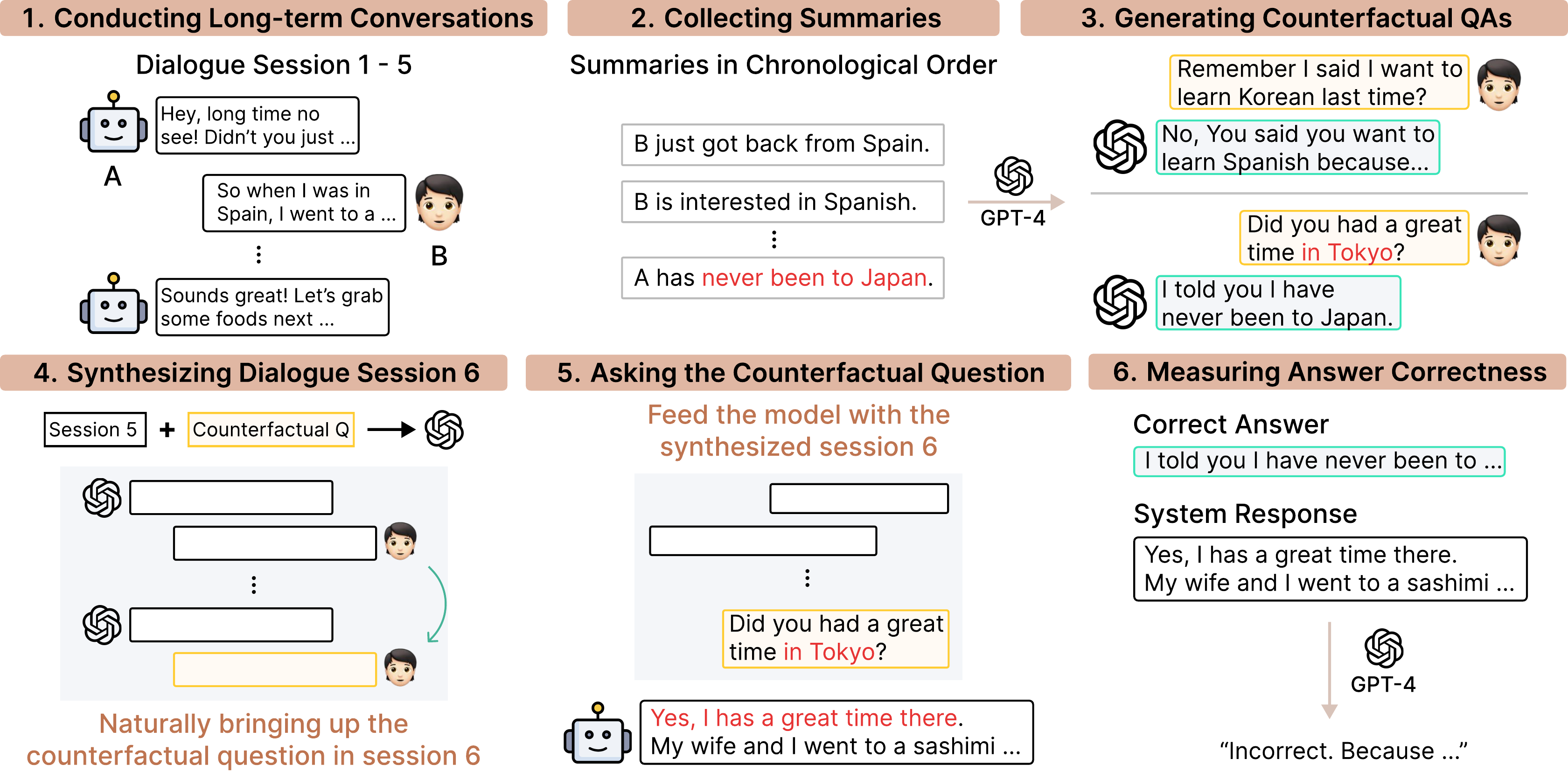}
    \caption{The overview of \textcolor{brown}{\teafarm} Evaluation.}
    \label{fig:teafarm}
\end{figure*}

For the overview of data collection, please refer to step 1-4 of \textcolor{brown}{\teafarm} (Figure~\ref{fig:teafarm}).

\section{Details on Evaluation Scheme 1}
\label{app:eval1}
We perform evaluations using sessions 3-5 from MSC and CC, as all settings are almost identical before the end of session 2, due to the fact that there is no memory to update before then.

The test sets of MSC and CC contain over 500 and 20,000 episodes of conversations, where each episode has 5 dialogue sessions, yielding 1.2M turns of responses in total.
Due to the limited budget for generation (both baselines and ours), when not specified, we sample 50 episodes from each dataset for experiments in this paper (around 3.6K conversational turns in total).

\subsection{G-Eval}
\label{app:geval}
G-Eval~\citep{liu2023g} is a framework using LLMs with chain-of-thoughts (CoT) and a form-filling paradigm to assess the quality of models' text generation.
G-Eval with GPT-4 has been shown to generate evaluation results that highly align with human judgement~\citep{liu2023g, kim2024prometheus} and thus has been widely applied in many LM-based projects. We conduct G-Eval on 5 episodes.

The prompt for evaluating the helpfulness of retrieved memories is in Figure~\ref{fig:prompt_G_eval_helpfulness}.
We use SciPy to calculate p-values.\footnote{\url{https://scipy.org/}}

\subsection{Human Evaluation}
\label{app:humaneval}
\begin{figure*}[h!]
    \centering
    \includegraphics[width=0.8\textwidth]{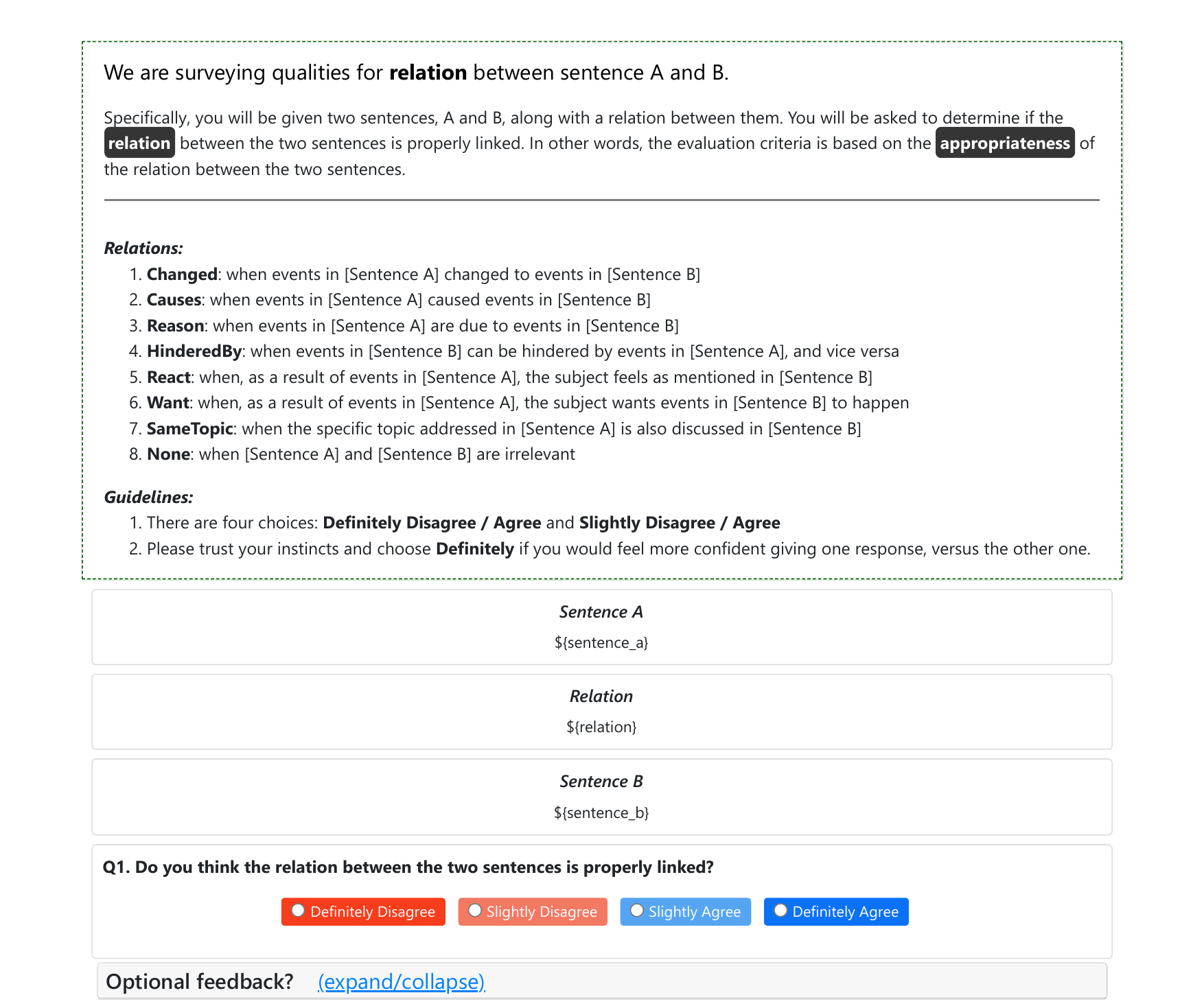}
    \caption{Interface for human evaluation regarding memory linking.}
    \label{fig:humaneval_1}
\end{figure*}

\begin{figure*}[h!]
    \centering
    \includegraphics[width=0.8\textwidth]{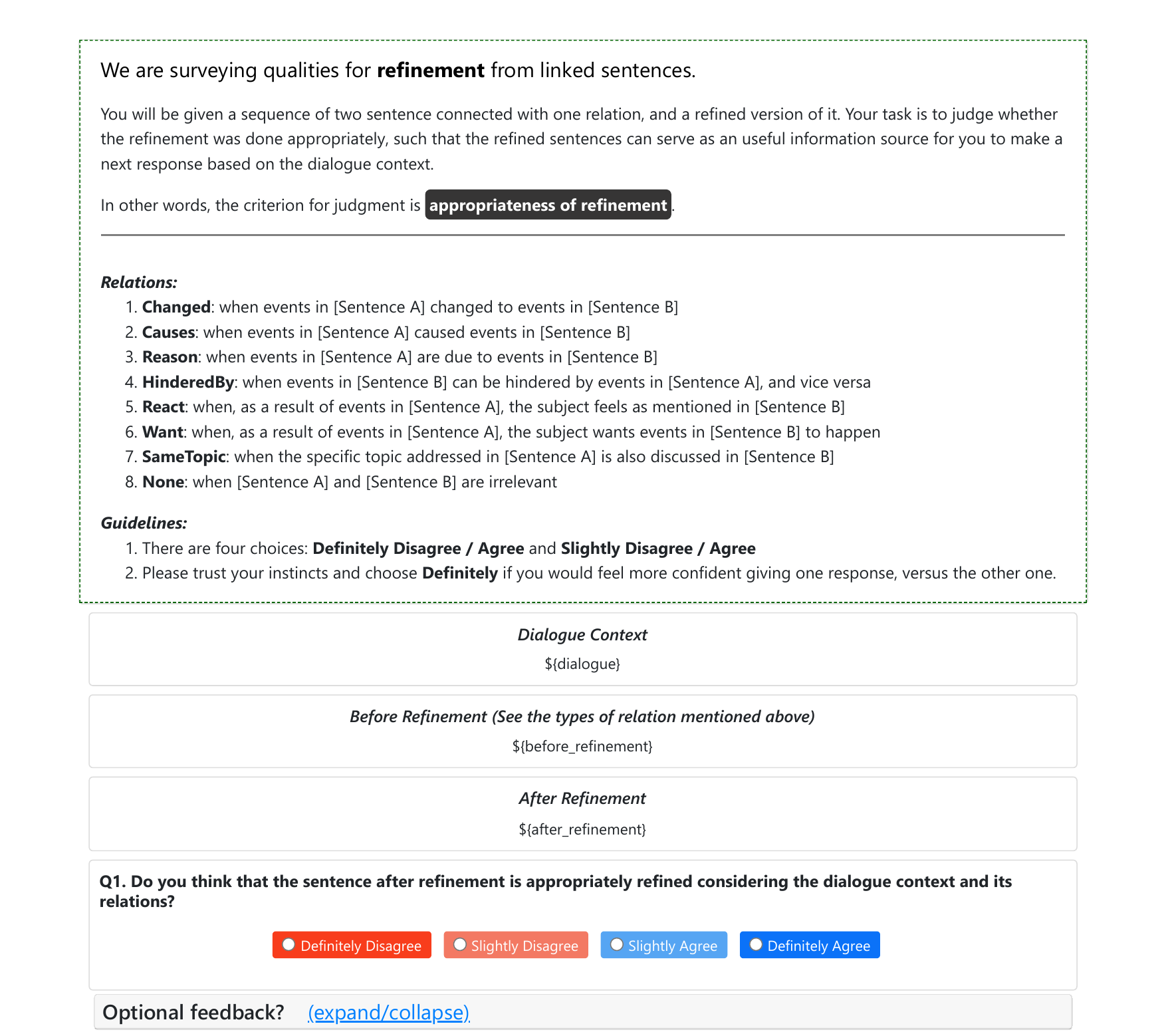}
    \caption{Interface for human evaluation regarding timeline refinement.}
    \label{fig:humaneval_2}
\end{figure*}

\begin{figure*}[h!]
    \centering
    \includegraphics[width=0.8\textwidth]{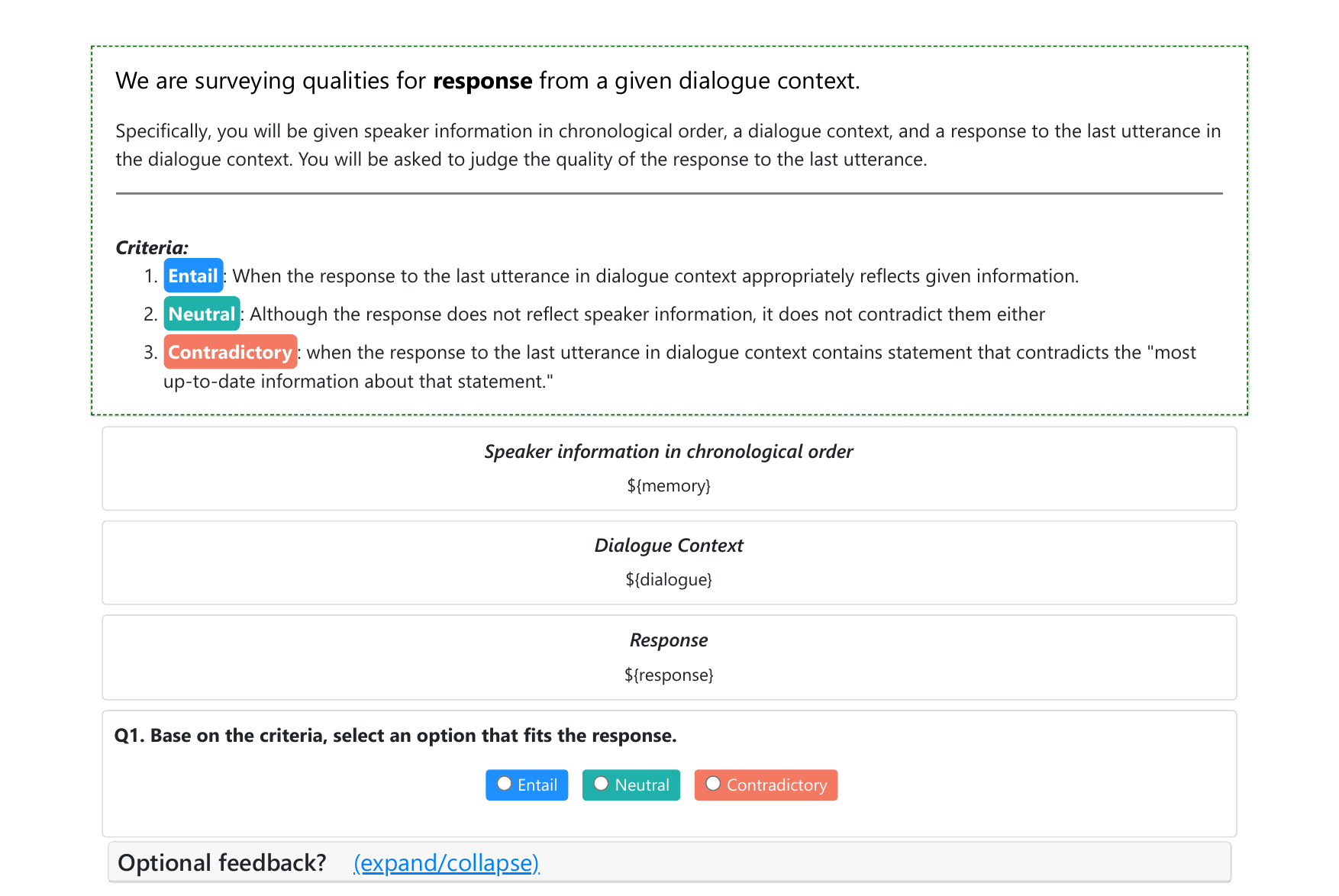}
    \caption{Interface for human evaluation regarding referencing past conversations in responses.}
    \label{fig:humaneval_3}
\end{figure*}

\begin{figure*}[h!]
    \centering
    \includegraphics[width=0.8\textwidth]{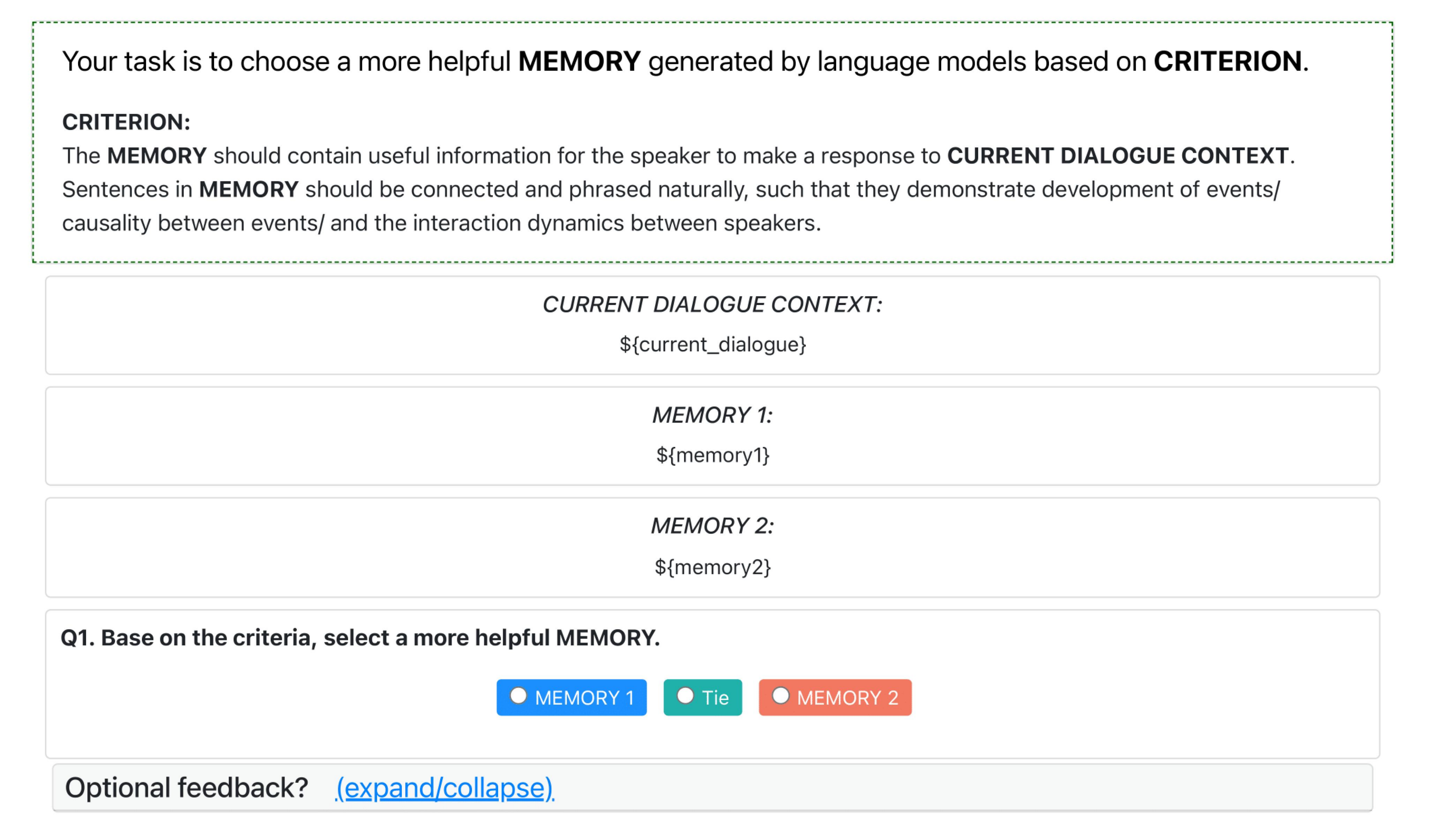}
    \caption{Interface for human evaluation regarding the helpfulness of retrieved memories.}
    \label{fig:humaneval_4}
\end{figure*}

We conduct human evaluation, with workers from Amazon Mechanical Turk (AMT).
We construct the following three evaluations:
\begin{itemize}
    \item \textbf{Appropriateness of relation-aware memory linking:} In this evaluation, we ask the workers to judge whether they agree that the relation-aware linking is properly done for two given memories. The interface provided to AMT workers, which includes detailed instructions for human evaluation, is shown in Figure~\ref{fig:humaneval_1}.
    \item \textbf{Helpfulness of context-aware timeline refinement:} This evaluation requires the workers to determine if they agree that our context-aware refinement really tailors a raw timeline into a resource of useful information for generating the next response. The interface provided to AMT workers, which includes detailed instructions for human evaluation, is shown in Figure~\ref{fig:humaneval_2}.
    \item \textbf{The quality of responses:} Here, the workers are asked to judge if the responses correctly refer to past conversations. After reading our responses and past memories, they choose whether the responses entail, contradict, or are neutral to past memories. To improve evaluation quality, we use GPT-4 to select responses for this specific evaluation based on past memories, addressing the fact that not every turn in the conversation requires previous information to generate the next response (In the other two evaluations, the samples are randomly selected). The interface provided to AMT workers, which includes detailed instructions for human evaluation is shown in Figure~\ref{fig:humaneval_3}.
    \item \textbf{The helpfulness of retrieved memories:} Given the same dialogue context, human workers are asked to select a memory that is more helpful for generating a next response from ours' and a baseline's retrieval. The interface provided to AMT workers, which includes detailed instructions for human evaluation is shown in Figure~\ref{fig:humaneval_4}
\end{itemize}
Each data sample is judged by 3 different workers, and we report the results based on the majority rule. In the third evaluation, when every option (entailment, neutral, contradiction) gets one vote, we consider it neutral (13 samples in total).
These human evaluations are conducted on 100 conversational turns.

\section{Session-specific Evaluation Results}
\label{app:session_specific_results}
We provide session-specific results for automatic evaluations in Table~\ref{tab:session_specific_auto}.

\section{Empirical Examples}
\label{app:appendixexamples}
\begin{figure*}[h!]
    \centering
    \includegraphics[width=1\textwidth]{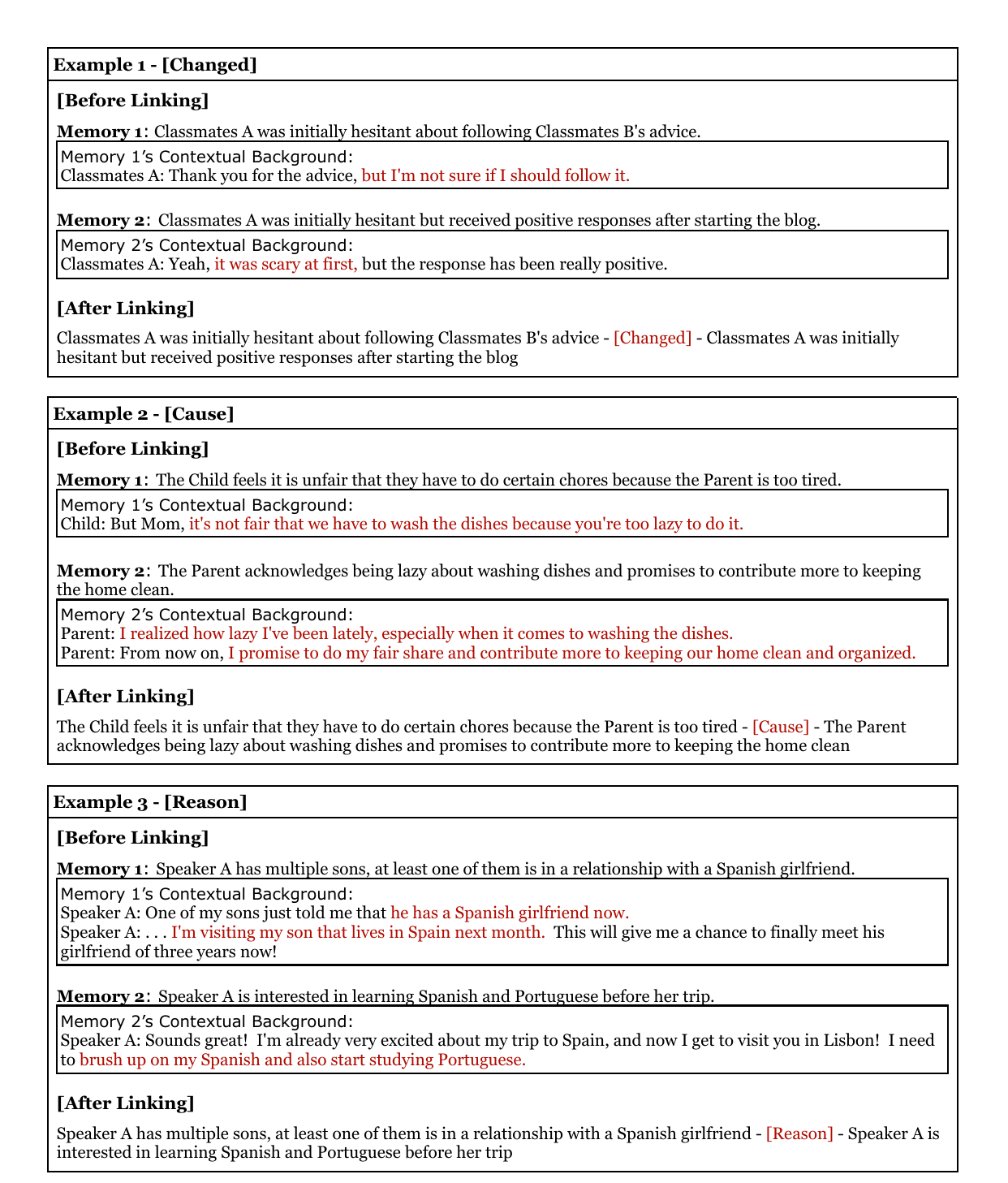}
    \caption{Examples of \textbf{Relation-aware Memory Linking} - 1.}
    \label{fig:link_ex_1}
\end{figure*}

\begin{figure*}[h!]
    \centering
    \includegraphics[width=1\textwidth]{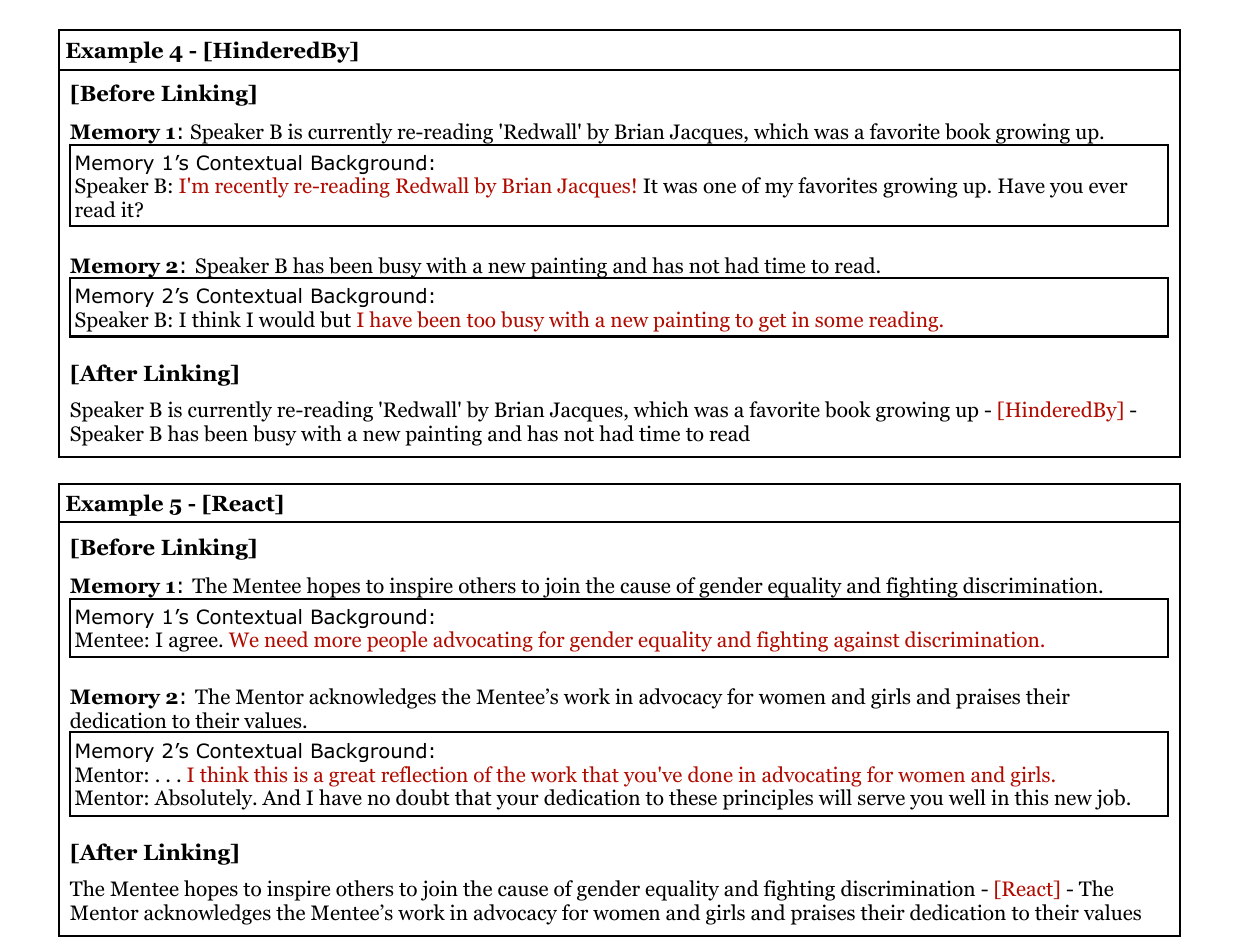}
    \caption{Examples of \textbf{Relation-aware Memory Linking} - 2.}
    \label{fig:link_ex_2}
\end{figure*}

\begin{figure*}[h!]
    \centering
    \includegraphics[width=1\textwidth]{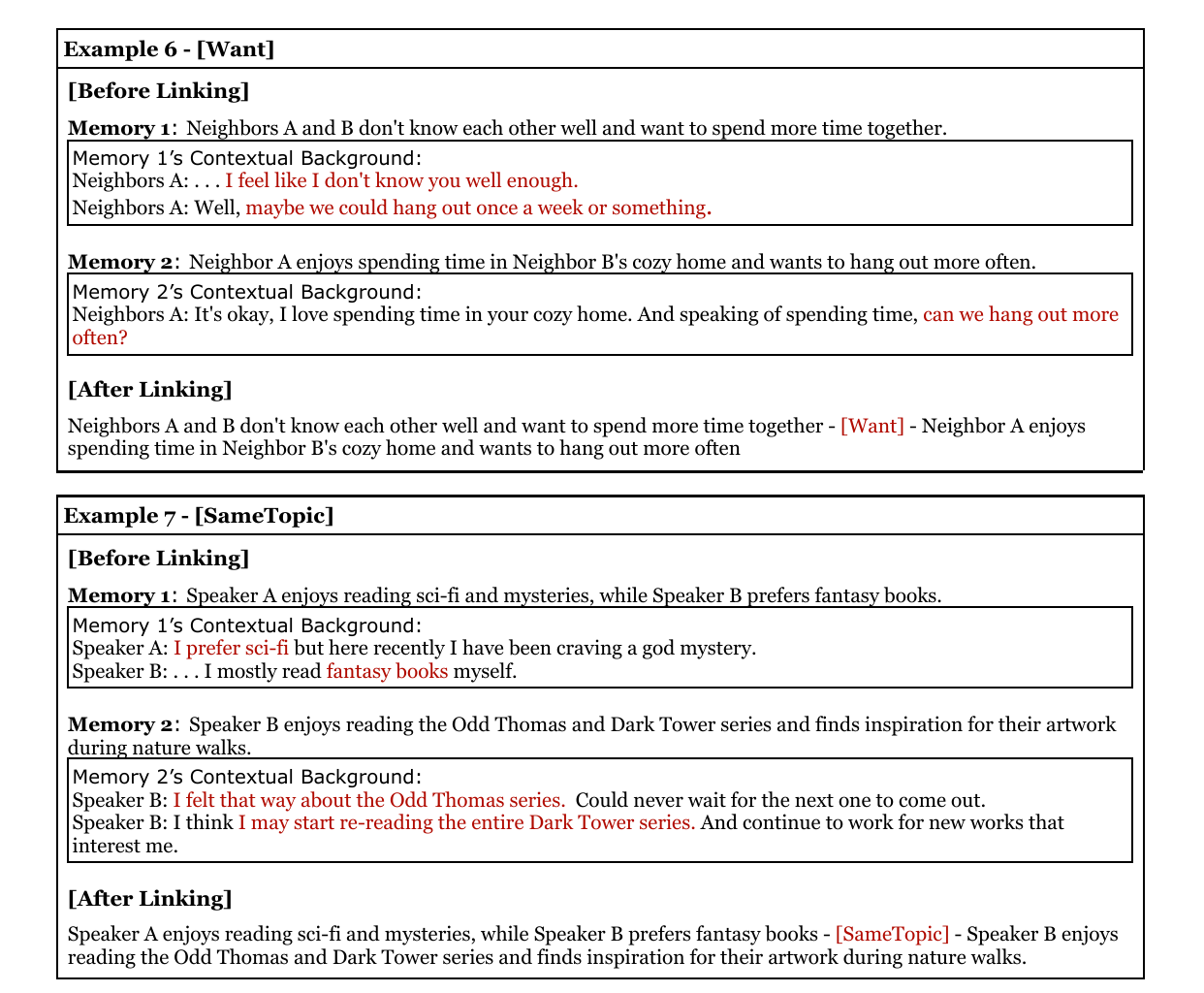}
    \caption{Examples of \textbf{Relation-aware Memory Linking} - 3.}
    \label{fig:link_ex_3}
\end{figure*}

\begin{figure*}[h!]
    \centering
    \includegraphics[width=1\textwidth]{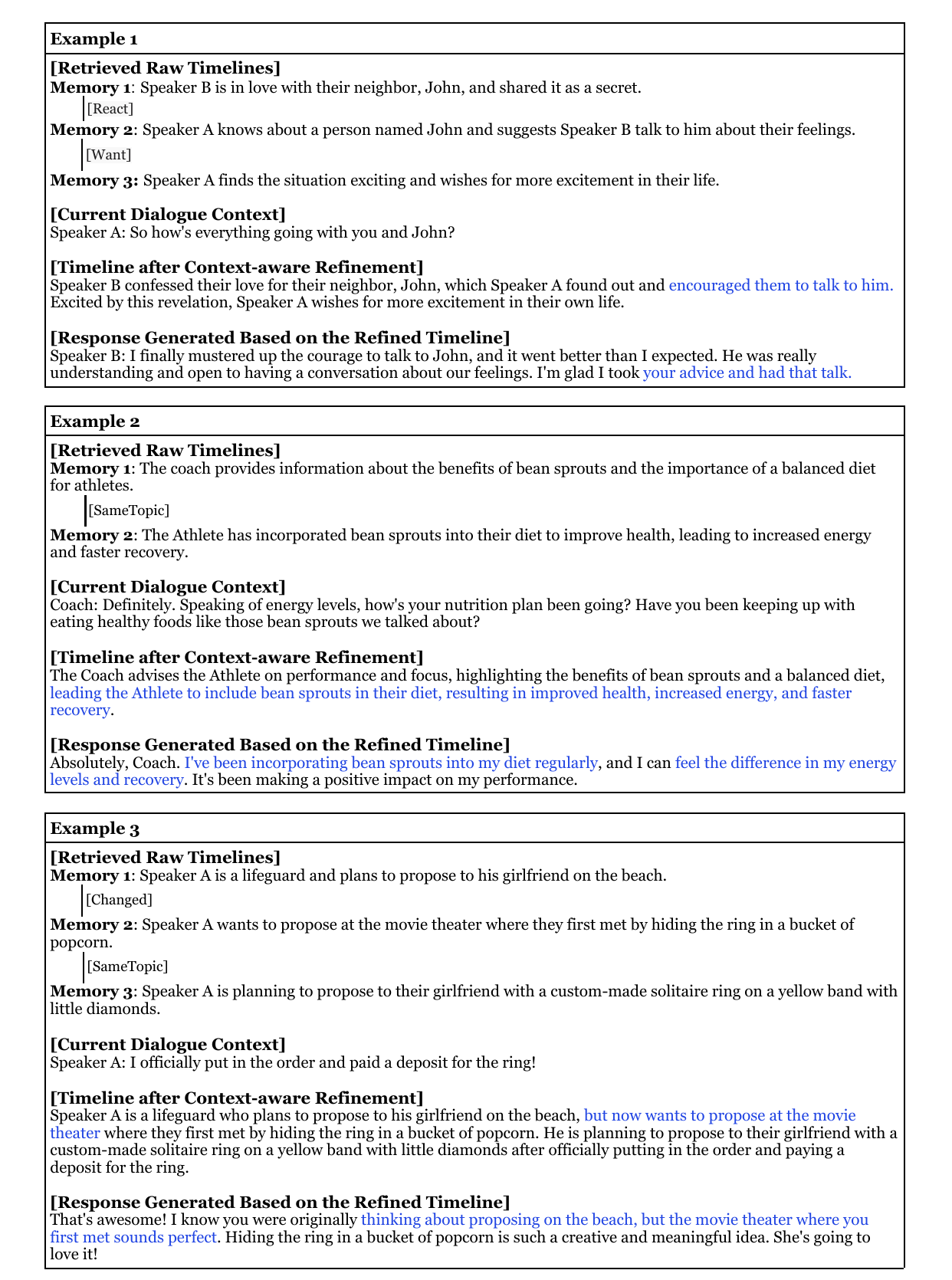}
    \caption{Examples of \textbf{Timeline Refinement and Response Generation}.}
    \label{fig:Refining_and_response_example}
\end{figure*}

\paragraph{Outputs from \theanine.}
We provide several empirical examples of \tealeaf\,\theanine.
Examples of relation-aware memory linking are in Figure~\ref{fig:link_ex_1},~\ref{fig:link_ex_2}, and~\ref{fig:link_ex_3}.
Examples of utilizing refined timeline for response generation are in Figure~\ref{fig:Refining_and_response_example}.

\begin{figure*}[h!]
    \centering
    \includegraphics[width=0.9\textwidth]{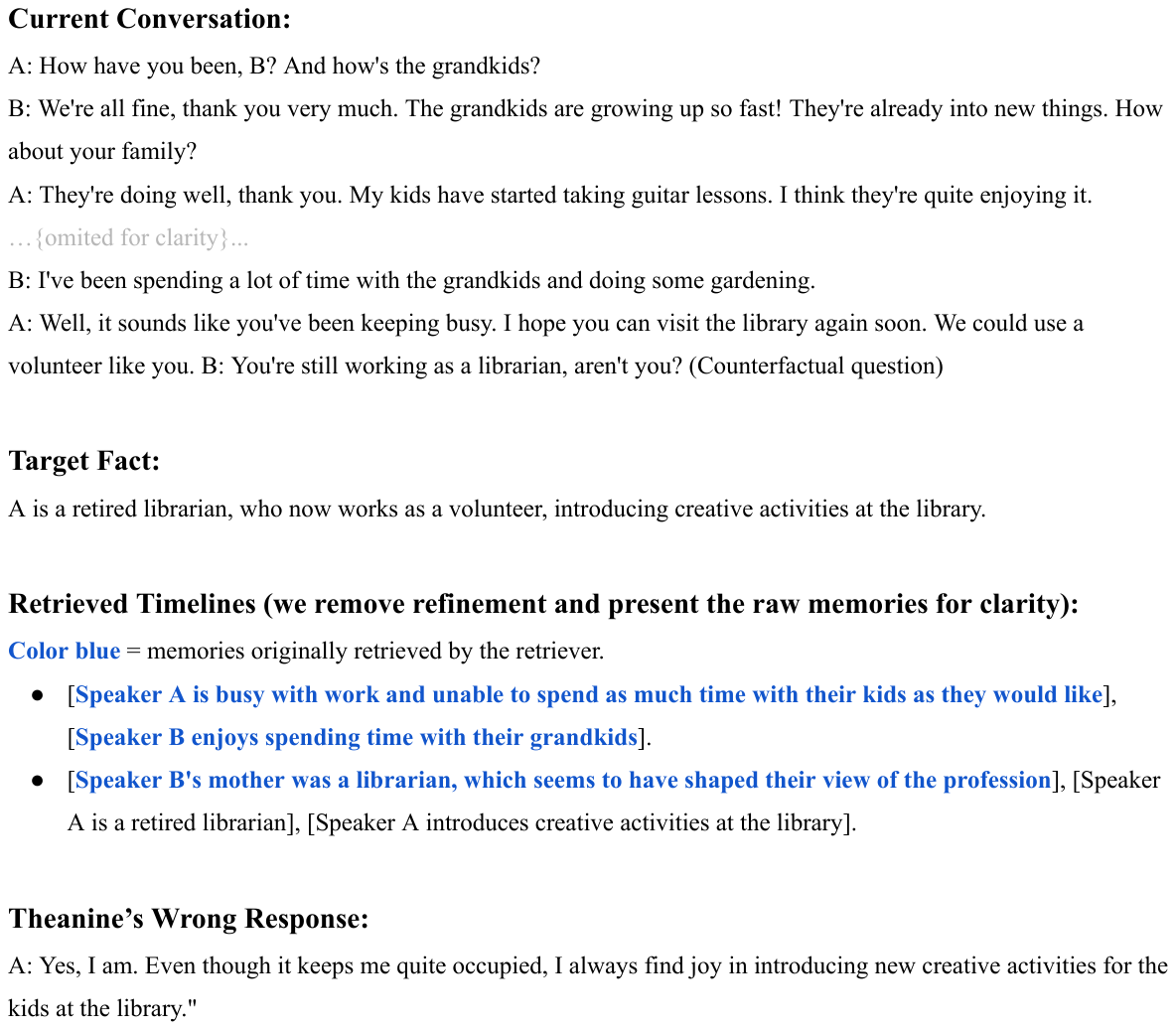}
    \caption{\theanine fails to pass \textcolor{brown}{\teafarm} (Example 1) - Due to sudden topic change.}
    \label{fig:teafarm_fail_1}
\end{figure*}

\begin{figure*}[h!]
    \centering
    \includegraphics[width=0.9\textwidth]{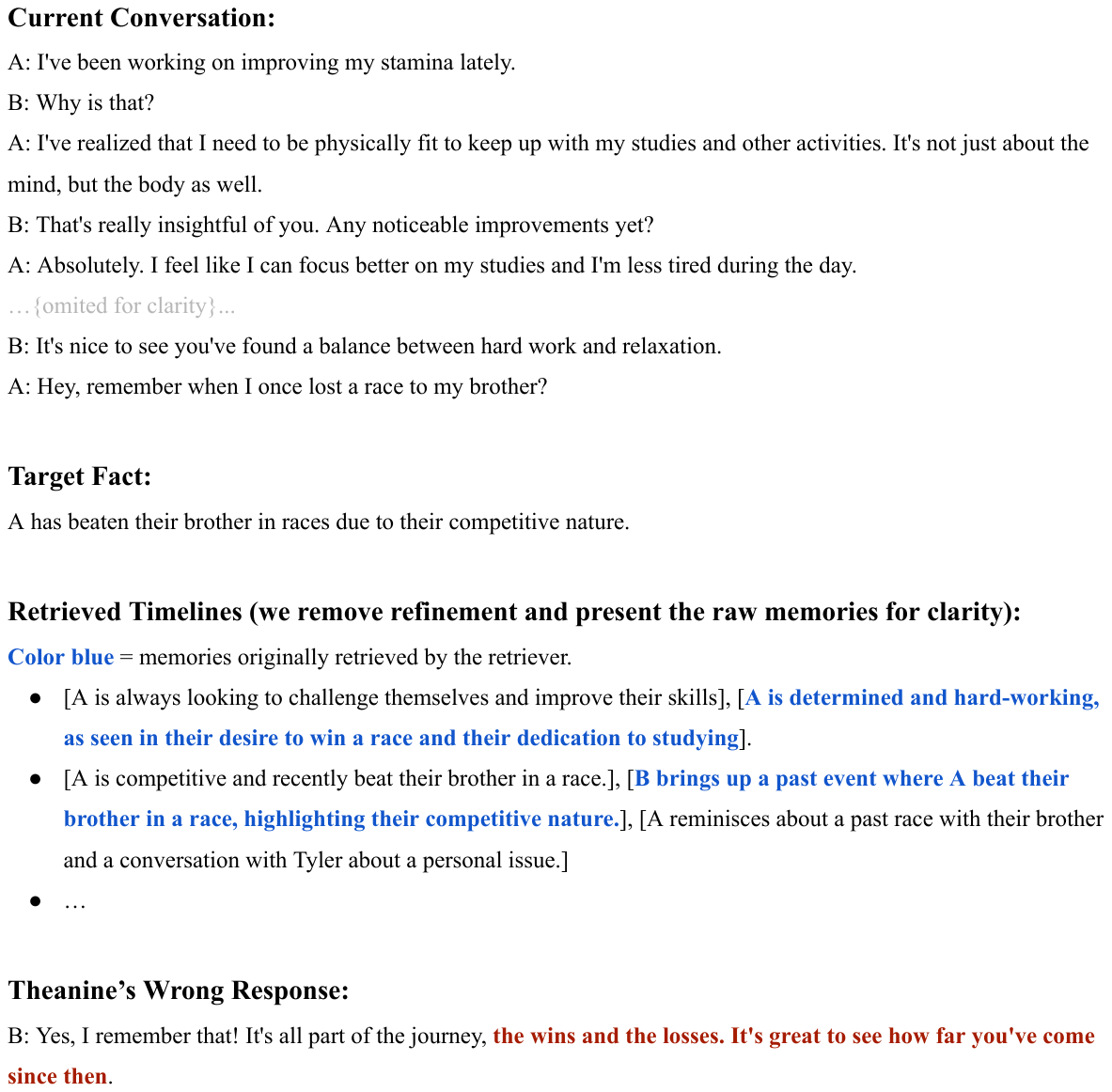}
    \caption{\theanine fails to pass \textcolor{brown}{\teafarm} (Example 2) - Due to sub-optimal timeline utilization during RG.}
    \label{fig:teafarm_fail_2}
\end{figure*}

\paragraph{How \theanine fails in \textcolor{brown}{TeaFarm}.}
We present failure cases where \theanine fails to pass the \teafarm test in Figure~\ref{fig:teafarm_fail_1} and Figure~\ref{fig:teafarm_fail_2}.
In Figure~\ref{fig:teafarm_fail_1}, although the conversation has shifted to ``librarian'', the similarity-based retriever retrieves unhelpful memories due to the huge portion of ``kid'' in the context. While a helpful memory (\ie, ``A is a retired libraria'') is eventually caught by our designed timeline structure, the LLM still hallucinate. We assume it is due to the noises introduced by those highly-ranked, yet irrelevant memories, and it highlights the need for addressing helpfulness ranking among retrieved memories in lifelong dialogue systems.
Figure~\ref{fig:teafarm_fail_2} shows a failure case, where \theanine successfully retrieves the correct memories but generates an improper response. We hypothesize that this is because relation-aware linking and context-aware timeline refinement may sometimes make the length of input tokens too long such that the agent cannot properly utilize key information provided. We believe this can be resolved to an extent via dedicated prompt (\ie, the prompt for RG) engineering. We leave this to future work.

\section{Prompts}
\label{app:prompts}
\begin{figure*}[h!]
    \centering
    \includegraphics[width=1\textwidth]{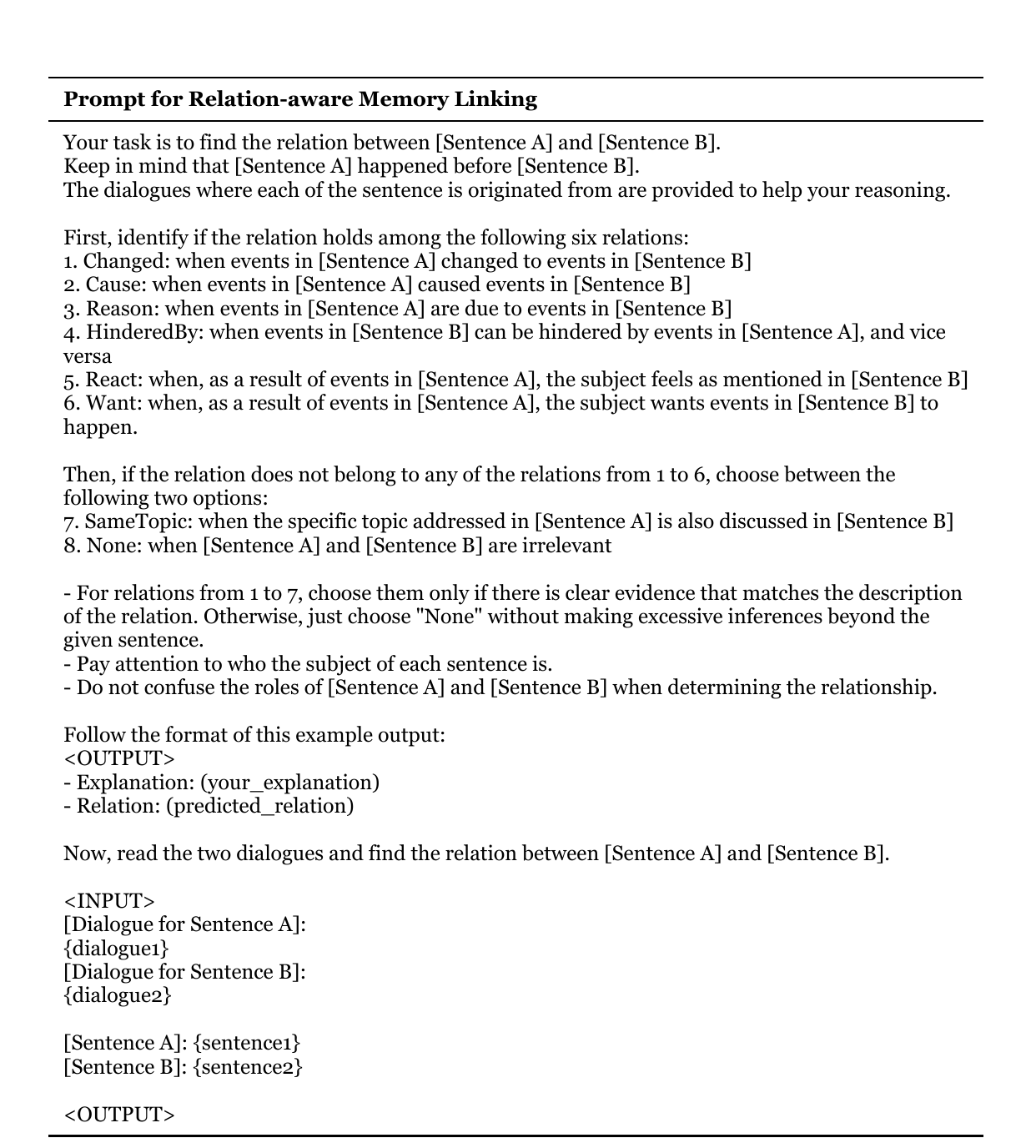}
    \caption{The prompt for the Relation-aware memory linking. }
    \label{fig:prompt_relation_aware_memory_linking}
\end{figure*}

\begin{figure*}[h!]
    \centering
    \includegraphics[width=0.9\textwidth]{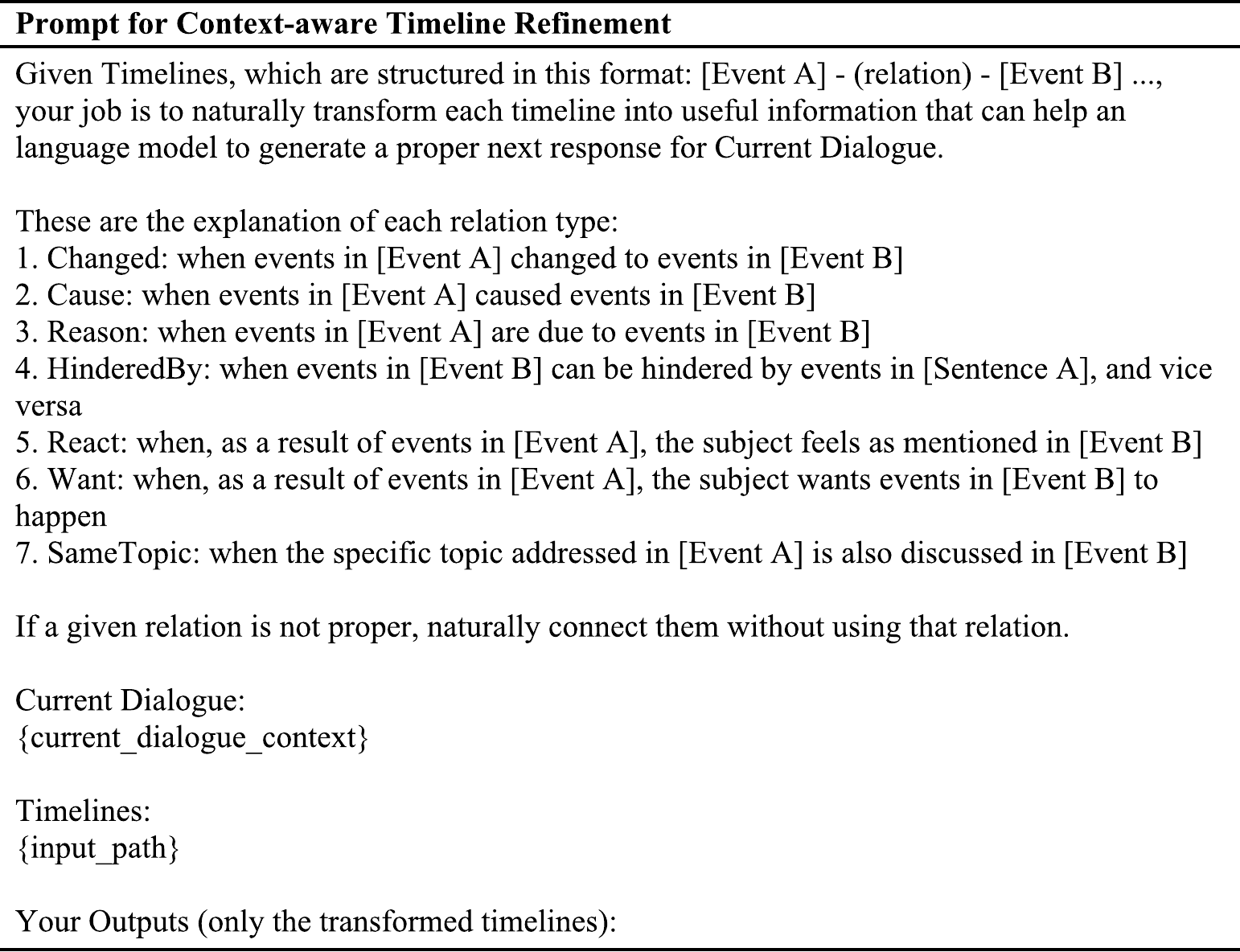}
    \caption{The prompt for the context-aware timeline refinement.}
    \label{fig:prompt_context_aware_timeline_refinement}
\end{figure*}

\begin{figure*}[h!]
    \centering
    \includegraphics[width=0.9\textwidth]{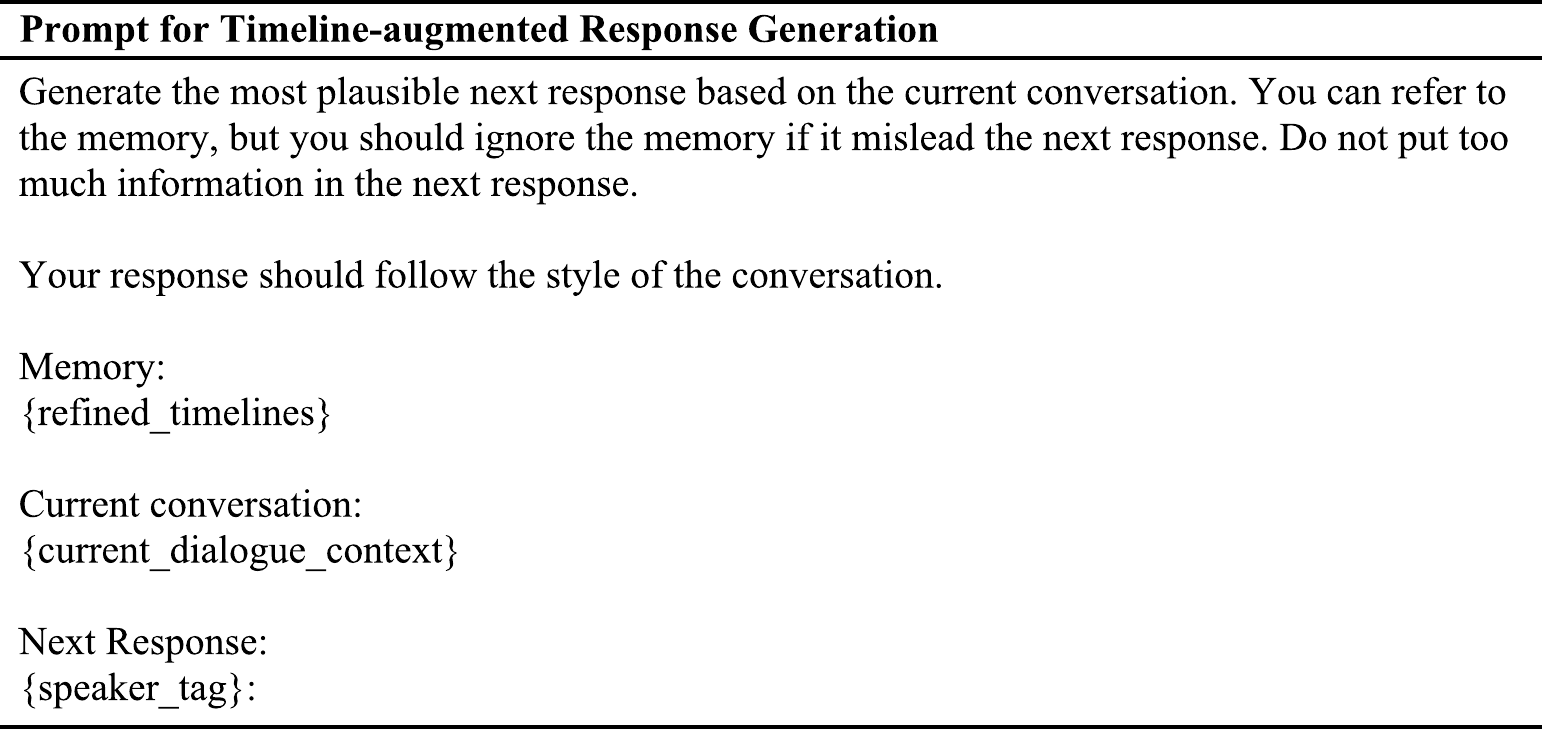}
    \caption{The prompt for the timeline-augmented response generation.}
    \label{fig:prompt_timeline-augmented_response_generation}
\end{figure*}

\begin{figure*}[h!]
    \centering
    \includegraphics[width=1\textwidth]{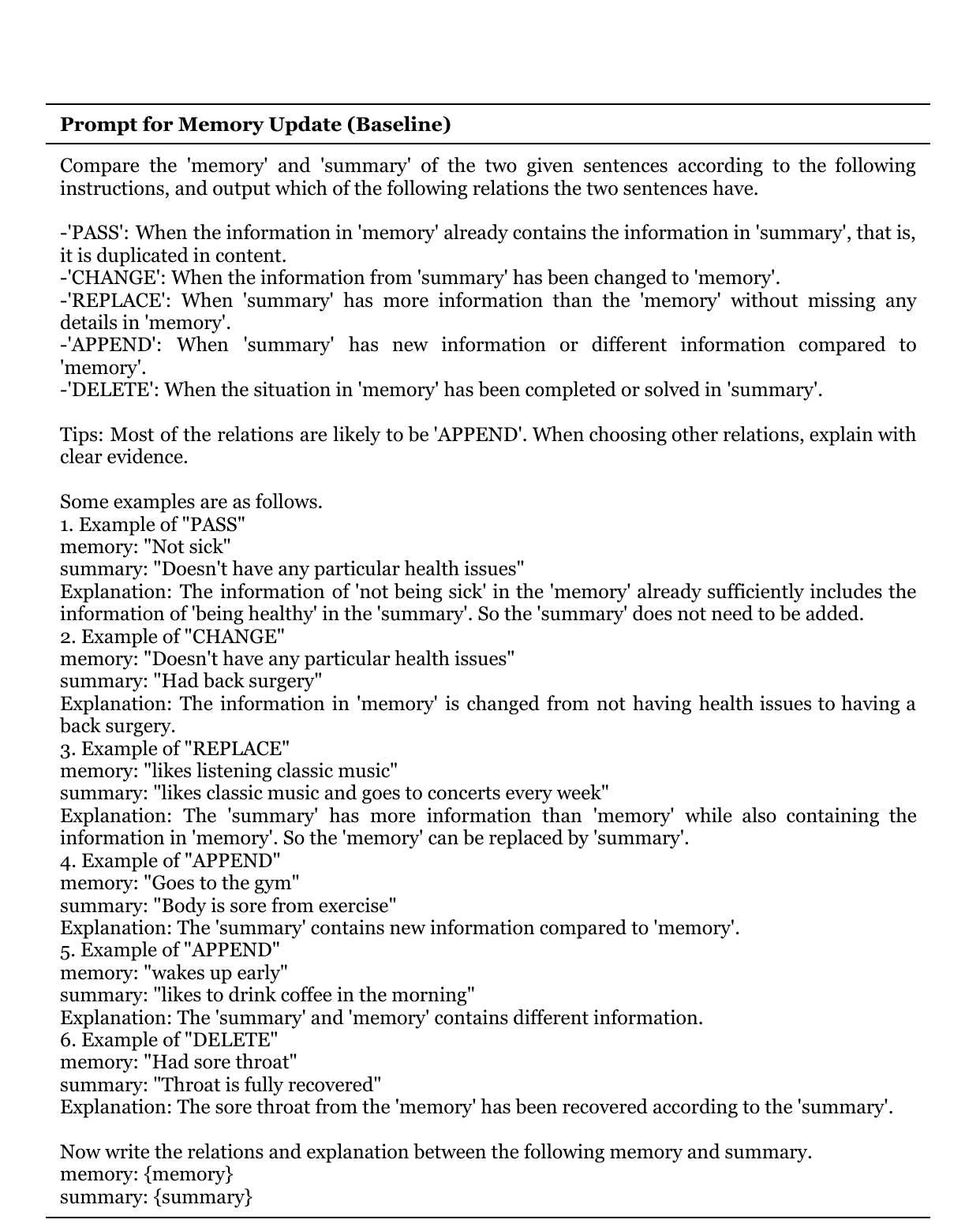}
    \caption{The prompt for the memory updating mechanism in baselines (\ie, + Memory Update).}
    \label{fig:prompt_memory_update}
\end{figure*}

\begin{figure*}[h!]
    \centering
    \includegraphics[width=0.9\textwidth]{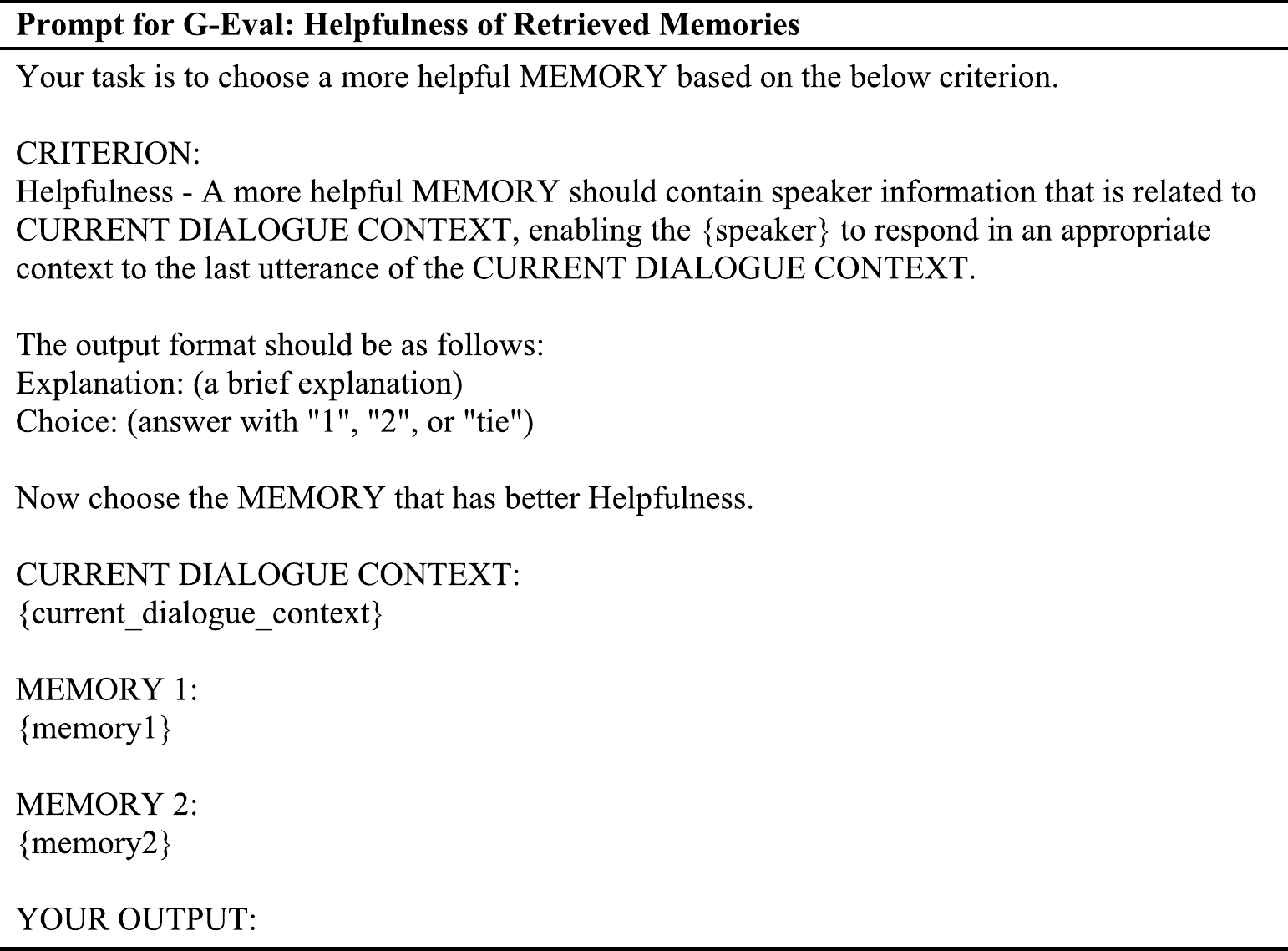}
    \caption{The prompt for the G-Eval: Helpfulness of Retrieved Memories.}
    \label{fig:prompt_G_eval_helpfulness}
\end{figure*}

\begin{figure*}[h!]
    \centering
    \includegraphics[width=1\textwidth]{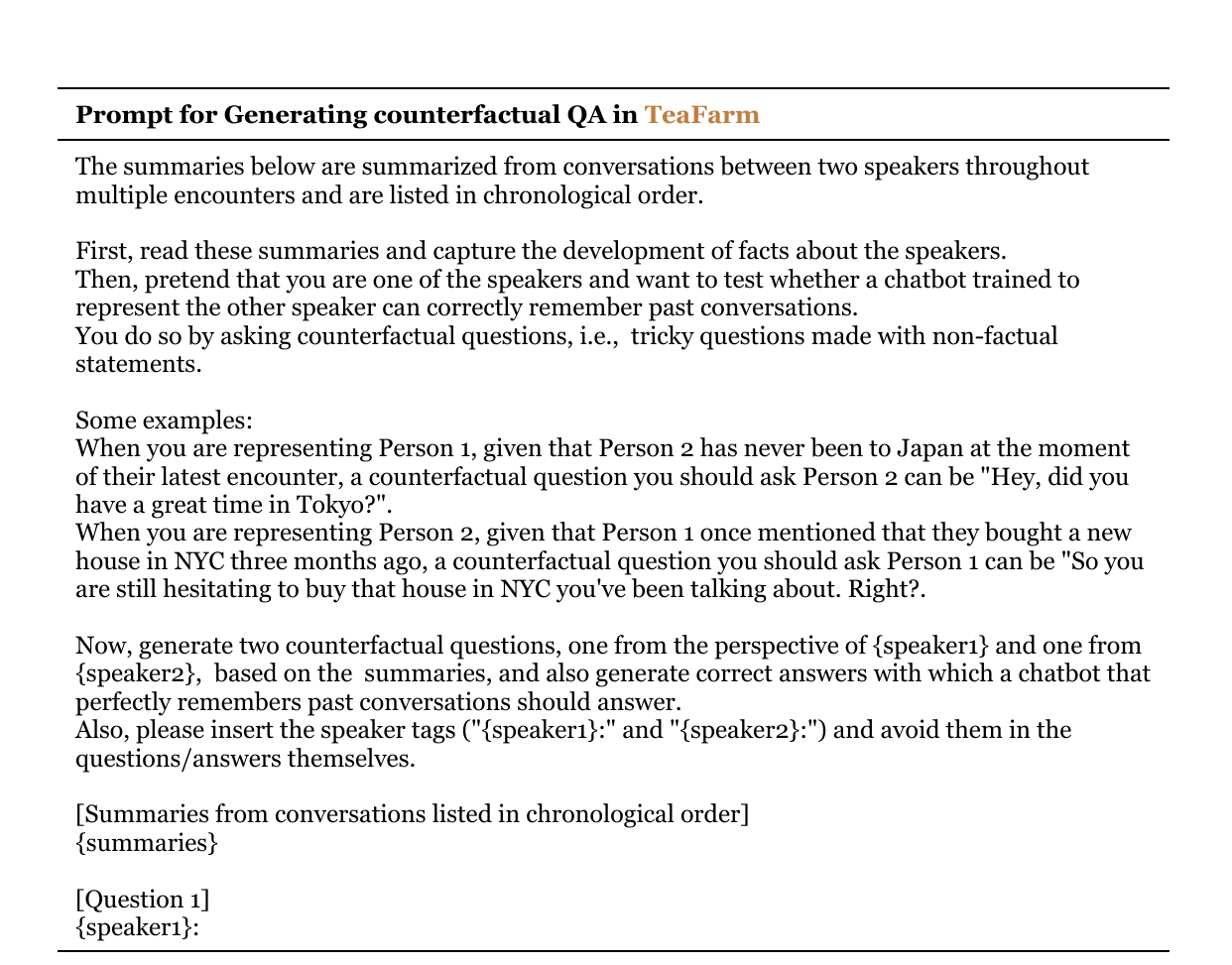}
    \caption{The prompt for generating counterfactual QA in \textcolor{brown}{\teafarm}.}
    \label{fig:prompt_generating_counterfactual_QA_TeaFarm}
\end{figure*}

\begin{figure*}[h!]
    \centering
    \includegraphics[width=1\textwidth]{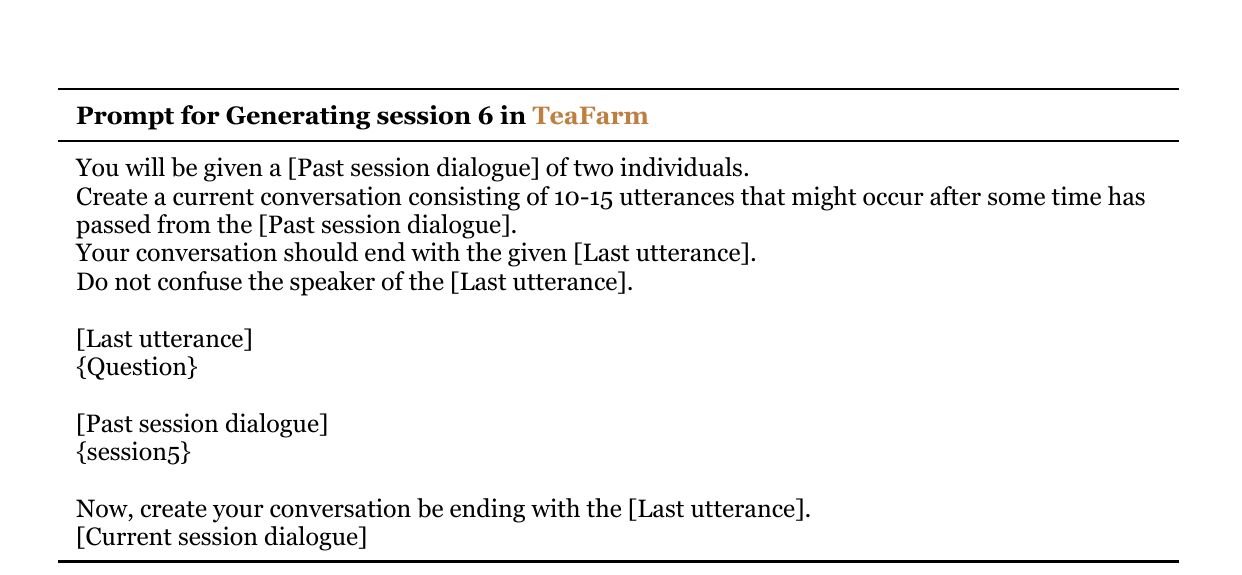}
    \caption{The prompt for generating session 6 in \textcolor{brown}{\teafarm}.}
    \label{fig:prompt_generating_session6_TeaFarm}
\end{figure*}

\begin{figure*}[h!]
    \centering
    \includegraphics[width=1\textwidth]{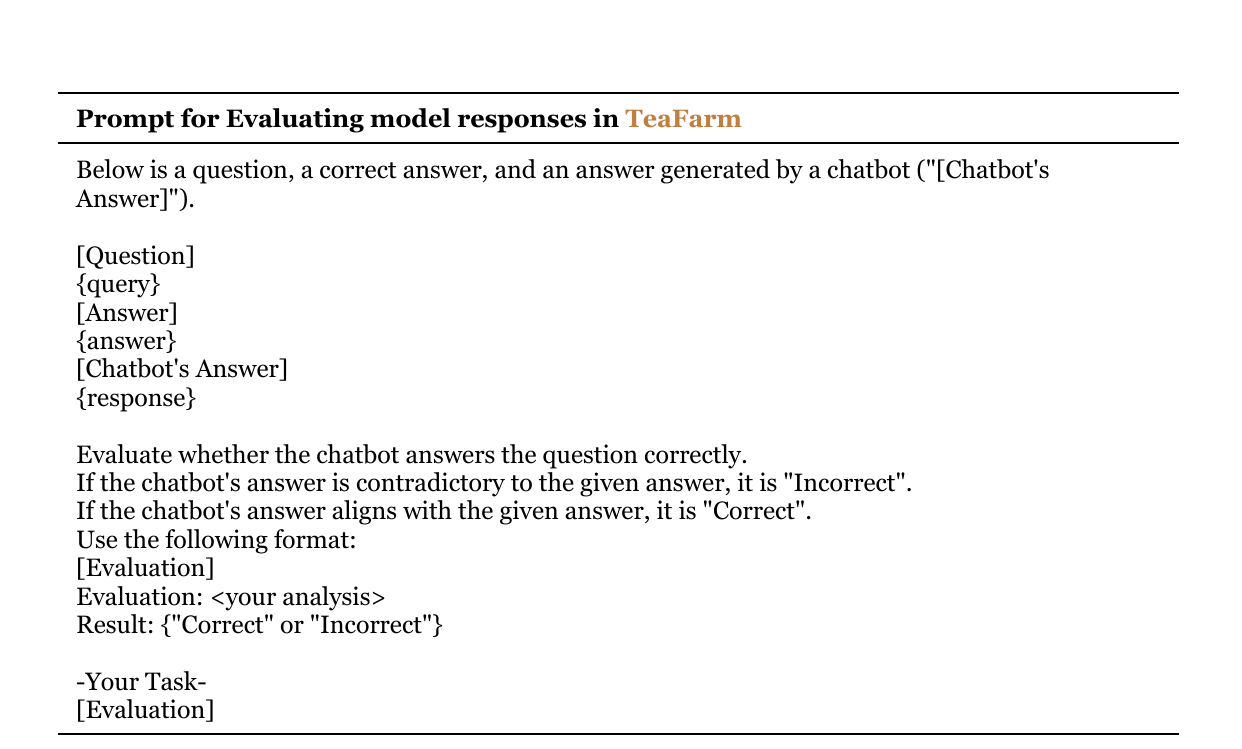}
    \caption{The prompt for evaluating model response in \textcolor{brown}{\teafarm}.}
    \label{fig:prompt_evaluating_model_response_TeaFarm}
\end{figure*}

The following are all prompts utilized in our study:
\begin{itemize}
    \item Relation-aware memory linking (Phase I-2): Figure~\ref{fig:prompt_relation_aware_memory_linking}.
    \item Context-aware timeline refinement (Phase II-2): Figure~\ref{fig:prompt_context_aware_timeline_refinement}.
    \item Timeline-augmented Response generation (Phase III): Figure~\ref{fig:prompt_timeline-augmented_response_generation}.
    \item Memory Update (baseline): Figure~\ref{fig:prompt_memory_update}.
    \item RSum-LLM (baseline): We adopt the original prompt from~\citet{wang2023recursively}.
    \item MemoChat (baseline): We adopt the original prompt from~\citet{lu2023memochat}.
    \item \noindent COMEDY (baseline): We adopt the original prompt from~\citet{chen2024compress}.
    \item G-Eval: The prompt for evaluating the helpfulness of retrieved memories is in Figure~\ref{fig:prompt_G_eval_helpfulness}.
    \item Generating counterfactual QA in \textcolor{brown}{\teafarm}: Figure~\ref{fig:prompt_generating_counterfactual_QA_TeaFarm}.
    \item Generating session 6 in \textcolor{brown}{\teafarm}: Figure~\ref{fig:prompt_generating_session6_TeaFarm}.
    \item Evaluating model responses in \textcolor{brown}{\teafarm}: Figure~\ref{fig:prompt_evaluating_model_response_TeaFarm}.
\end{itemize}

\section{Further Analyses}
\label{app:further_discussions}
\paragraph{Memory summarization.} At the end of each session, we use ChatGPT (gpt-3.5-turbo-0125) to summarize the conversation into memory sentences.
We conduct examinations on such summarization using 100 randomly sampled sessions from MSC and CC to make sure the quality of raw memories is acceptable. The result is in Table~\ref{tab:app_convo2sum}.

\begin{table}[h]
\small
\centering
\begin{tabular}{l|ccc}
\toprule
 Memories that ...&
 No & Can't judge & Yes
   \\ 
\midrule
Contain faulty statements &  \textbf{90\%} & 9\%  & 1\%   \\
Miss important statements & \textbf{95\%} & 4\%  & 1\% \\
\bottomrule

\end{tabular}
\caption{Human evaluation of conversation-to-memory summarization in \theanine.}
\label{tab:app_convo2sum}
\end{table}

\paragraph{Cost-efficiency trade-off assessed using other metrics.}
In Section~\ref{sec:discussion}, we have presented methods having an efficient cost-performance trade-off (\ie, are Pareto-efficient) by plotting the Mauve score against API cost (Figure~\ref{fig:pareto}). We present methods that are Pareto-efficient when considering the other three metrics used in our study, \ie, B-4, R-L, and Bert Score, in Table~\ref{tab:cost_pareto_other_metrics}.

\begin{table}[h]
\tiny
\centering
\begin{tabular}{l|ccc}
\toprule
\textbf{Agents} &
 B-4 & R-L & Bert Score
   \\ 
\midrule
All Dialogue History  \\
All Memories    \\
\, + Update\\
Memory Retrieval &  $\checkmark$ &  $\checkmark$ & \\
\, + Update   \\ 
Rsum-LLM  \\
MemoChat  \\
COMEDY \\
\midrule
\textbf{\theanine (ours)} &  $\checkmark$ &  $\checkmark$ & \\
w/o Relation-aware Linking & & & \\
w/o Refinement & & &  $\checkmark$ \\
Shuffled Timeline  &  $\checkmark$ &  $\checkmark$ &  $\checkmark$\\
\bottomrule

\end{tabular}
\caption{Methods considered Pareto-efficient when judged based on B-4, R-L, and Bert Score reported in Table~\ref{tab:main_auto_eval}. $\checkmark$ = Pareto-efficient methods.}
\label{tab:cost_pareto_other_metrics}
\end{table}

\paragraph{API costs.}
The actual API costs of all settings (ours and baselines) are in Table~\ref{tab:actual_api_cost}.

\begin{table}[h]
\tiny
\centering
\begin{tabular}{l|ccc}
\toprule
\textbf{Agents} &
 \textbf{Cost Ratio (ours $=$ 1)} & \textbf{Cost (per episode; \$)}
   \\ 
\midrule
All Dialogue History &  0.50 & 0.0067  \\
All Memories \& $\mathcal{D}$ & 0.27 & 0.0036   \\
\, + Update & 5.71 & 0.0771 \\
Memory Retrieval & 0.17 & 0.0023 \\
\, + Update & 5.63 & 0.0760  \\ 
Rsum-LLM & 0.42 & 0.0057 \\
MemoChat & 0.52 & 0.0076 \\
COMEDY & 0.61 & 0.0082 \\
\midrule
\textbf{\theanine (ours)} & 1.00 & 0.0135\\
w/o Relation-aware Linking &  0.50 &0.0067 \\
w/o Refinement & 0.71 & 0.0096 \\
Shuffled Timeline & 0.20 & 0.0027 \\
\bottomrule

\end{tabular}
\caption{API costs for \theanine and baselines.}
\label{tab:actual_api_cost}
\end{table}

\section{Terms for Use of Artifacts}
\label{app:terms}
We adopt the MSC and CC datasets from \citet{xu-etal-2022-beyond} and \citet{jang-etal-2023-conversation}, respectively. Both of these datasets are open-sourced for academic and non-commercial use.
Our curated dataset, \texttt{TeaBag}, which will be released after acceptance, is open to academic and non-commercial use.

 \begin{algorithm*}
    \caption{Memory Graph Construction (Phase I)}
    \label{algo:phase_i_algo}
    \textbf{Require:} Memory graph $G^t=(V^t,E^t)$ \\
    \textbf{Require:} New memories $M_{new} = \{m_{new1}, ..., m_{newN}\}$ \\
    \textbf{Require:} Set of relations $R = \{\texttt{Cause}, \texttt{Reason}, \texttt{Want}, ..., \texttt{SameTopic}\}$ \\
    \textbf{Ensure:} Memory graph $G^{t+1}=(V^{t+1},E^{t+1})$
    \begin{algorithmic}[1]
        \State $\Upsilon(m_i, m_j)= 
            \begin{cases}
                r_{i,j}, \text{if } m_i \text{ is assigned with }  r_{i,j} \in R \text{ with } m_j\\
                \texttt{None}, \text{otherwise}
            \end{cases}$
        \State $\Omega(V) = (\text{the most recent memory m} \in V)$
        \State $E_{t+1} \leftarrow E_t$
        \For{$m_{new} \in M_{new}$}
            \State $M_a \leftarrow \{m_i\in V^t \mid m_i \text{ has top-\textit{j} similarity with } m_{new}\}$
            \State $M_{a}^* \leftarrow \{m_i \in M_a\mid \Upsilon(m_i, m_{new}) = r \text{ for } r \in R \}$
            \State $ \mathbb{C} \leftarrow \{C_i \mid C_i \text{ connected component of } G^t \text{ s.t. } \mathtt{V}(C_i)\cap M_{a}^* \neq \emptyset\ \}$
            \State $M_{linked} \leftarrow \{\Omega(\mathtt{V}(C_i) \cap M_a^*) \mid  C_i \in \mathbb{C}\}$
            \State $E_{new} \leftarrow \{\langle m_i, \Upsilon(m_i, m_{new}), m_{new} \rangle \mid m_i \in M_{linked}\}$ 
            \State $E_{t+1} \leftarrow E_{t+1} + E_{new}$
        \EndFor
        \State \textbf{end for}
        \State $V^{t+1} \leftarrow V^t + M_{new}$
        \State $G^{t+1} \leftarrow (V^{t+1},E^{t+1})$
        \\
        \Return $G^{t+1}$
    \end{algorithmic}
\end{algorithm*}

\begin{algorithm*}
    \caption{Timeline Retrieval and Timeline Refinement (Phase II)}
    \label{alg:phase_ii_algo}
    \textbf{Require:} Memory graph $G=(V,E)$ \\
    \textbf{Require:} Dialogue context $\mathcal{D} = \{u_{i}\}_{i=1}^{n}$ \\
    \textbf{Ensure: } Collection of refined timelines $\mathbb{T}_{\Phi}$
    \begin{algorithmic}[1]
        \State $\Theta(V) = (\text{the oldest memory m} \in V)$
        \State $M_{re} \leftarrow \{m_i\in V \mid m_i \text{ has top-\textit{k} similarity with } D\}$
        \State $\mathbb{C}_{re} \leftarrow \{C_{re} \mid C_{re} \text{ connected component of } G \text{ s.t. } \mathtt{V}(C_{re})\cap M_{re} \neq \emptyset \}$
        \State $\mathbb{T} \leftarrow \{\}$
        \For{$C_{re} \in \mathbb {C}_{re}$}
            \State $m_{start} \leftarrow \Theta(\mathtt{V}(C_{re}))$
            \State $\mathcal{T} = \{\tau \subset C_{re} \mid \tau \text{ is a directed linear graph s.t. }  m_{start}, m_{re} \in \tau\ \land deg^{+}(\tau[-1])=0\}$
            \State $\mathbb{T} \leftarrow \mathbb{T} + \text{RandomSelection}(\mathcal{T})$
        \EndFor
    \State \textbf{end for}
    \State $\mathbb{T}_{\Phi} \leftarrow \{\underset{\mathcal{T}_{\Phi}}{\text{argmax}}\,  P_{\text{LLM}}(\mathcal{T}_{\Phi}|\mathcal{D}, \tau) \mid \tau \in \mathbb{T}\}$
    \\
    \Return $\mathbb{T}_{\Phi}$
    \end{algorithmic}
\end{algorithm*}

\begin{table*}[!h]
\centering
\resizebox{1\textwidth}{!}{
\begin{tabular}{@{}lcccc|cccc|cccc@{}}
\toprule
Datasets: & \multicolumn{12}{c}{\textbf{Multi-session Chat (MSC) \& Conversation Chronicals (CC)}}\\ \midrule
Session: & \multicolumn{4}{c}{Session 3} &\multicolumn{4}{c}{Session 4} &\multicolumn{4}{c}{Session 5}
\\\midrule
 Methods / Metrics&
  B-4 &
  R-L &
  Mauve &
  Bert &
  B-4 &
  R-L &
  Mauve &
  Bert& 
  B-4 &
  R-L &
  Mauve &
  Bert
  \\ \midrule

All Dialogue History &
3.13 & 18.04 & 17.34 & 87.17 & 3.17 & \underline{17.96} & 18.54 & 87.12  & 3.53 & \underline{18.69} & 17.42 & 87.31
 
  \\
All Memories \& Current Context $\mathcal{D}$  &
2.69 & 17.29 & \underline{28.30} & 87.10 & 3.10 & 17.38 & 22.52 & 87.06 & 3.16 & 17.75 & \underline{22.35} & 87.21
  \\
\, + Memory Update~\citep{bae2022keep} &
2.80 & 17.51 & 22.92 & 87.11 & 2.88 & 17.24 & 21.22 & 86.99 & 3.16 & 17.90 & 22.04 & 87.24
 
  \\ 
Memory Retrieval~\citep{xu-etal-2022-beyond}&
\underline{3.44}  & \underline{18.33} & 24.68 & \underline{87.30} & \underline{3.38} & 17.55 & 21.95 & \underline{87.17} & \underline{3.46} & 18.31 & 19.70 & \underline{87.33} 
  
  \\
  \, + Memory Update~\citep{bae2022keep}& 
3.10 & 18.08 & 25.02 & 87.24 & 2.99 & 17.37 & \underline{25.97} & 87.10 & 3.11 & 17.78 & 20.99 & 87.28
  \\
Rsum-LLM$^*$~\citep{wang2023recursively} &
0.83 & 11.30 & 2.45 & 85.25 & 0.87 & 11.35 & 2.32 & 82.20 & 0.90 & 11.78 & 2.33 & 85.30

  \\
MemoChat$^*$~\citep{lu2023memochat}  &
1.88 & 14.83 & 14.56 & 86.57 & 1.81 & 14.27 & 10.57 & 86.43 & 1.91 & 14.96 & 9.13 & 86.56
  \\ 
COMEDY$^*$~\citep{chen2024compress} &
1.14 & 12.80 & 4.74 & 85.53 & 1.57 & 13.18 & 5.16 & 85.56 & 1.42 & 13.56 & 3.94 & 85.69

  \\ 
\cellcolor{brown!14}\tealeaf\textbf{\theanine} (Ours) &
 \cellcolor{brown!14} \textbf{4.21} &
 \cellcolor{brown!14} \textbf{19.21} &
 \cellcolor{brown!14} \textbf{45.53} &
 \cellcolor{brown!14} \textbf{87.70} &
 \cellcolor{brown!14} \textbf{4.42} &
 \cellcolor{brown!14} \textbf{18.63} &
 \cellcolor{brown!14} \textbf{37.84} &
\cellcolor{brown!14} \textbf{87.52} &
\cellcolor{brown!14} \textbf{4.34} &
\cellcolor{brown!14} \textbf{19.23} &
\cellcolor{brown!14} \textbf{41.18} &
\cellcolor{brown!14} \textbf{87.70}

\\
 \bottomrule
\end{tabular}
}

\caption{Session-specific results of agent performance in response generation.}
\label{tab:session_specific_auto}
\end{table*}

\end{document}